\documentclass[twoside]{article}

\usepackage{times}  
\usepackage{helvet}  
\usepackage{courier}  
\usepackage[hyphens]{url}  
\usepackage{graphicx} 
\usepackage{algorithm}
\usepackage{algorithmic}

\usepackage{url}            
\usepackage{booktabs}       
\usepackage{amsfonts}       
\usepackage{nicefrac}       
\usepackage{microtype}      
\usepackage{xcolor}         
\newcommand{\R}{\mathbb{R}}
\usepackage{enumitem}
\usepackage{amsmath}
\usepackage{subcaption}
\usepackage{caption}
\usepackage{amsthm}
\usepackage{algorithm}
\usepackage{algorithmic}
\usepackage{multirow}
\usepackage{graphicx}
\usepackage{multicol}
\usepackage{lipsum}
\usepackage{arydshln}
\usepackage[title]{appendix}
\usepackage[colorlinks,citecolor=blue]{hyperref}

\DeclareMathOperator*{\argmin}{arg\,min}
\DeclareMathOperator{\supp}{supp}
\DeclareMathOperator{\diag}{Diag}

\newtheorem{theorem}{Theorem}[section]
\newtheorem{proposition}[theorem]{Proposition}
\newtheorem{lemma}[theorem]{Lemma}

\newtheorem{fact}[theorem]{Fact}

\newcommand{\modelname}{\texttt{FALCON}}

\usepackage[accepted]{aistats2024}

\usepackage[round]{natbib}

\begin{document}

%

%

\twocolumn[

\aistatstitle{FALCON: FLOP-Aware Combinatorial Optimization for Neural Network Pruning}

\aistatsauthor{ Xiang Meng \And Wenyu Chen \And  Riade Benbaki \And Rahul Mazumder }

\aistatsaddress{ MIT \And  MIT \And MIT \And MIT} ]

\begin{abstract}
  The increasing computational demands of modern neural networks present deployment challenges on resource-constrained devices. Network pruning offers a solution to reduce model size and computational cost while maintaining performance. However, most current pruning methods focus primarily on improving sparsity by reducing the number of nonzero parameters, often neglecting other deployment costs such as inference time, which are closely related to the number of floating-point operations (FLOPs). In this paper, we propose \modelname, a novel combinatorial-optimization-based framework for network pruning that jointly takes into account model accuracy (fidelity), FLOPs, and sparsity constraints. A main building block of our approach is an integer linear program (ILP) that simultaneously handles FLOP and sparsity constraints. We present a novel algorithm to approximately solve the ILP. We propose a novel first-order method for our optimization framework which makes use of our ILP solver. Using problem structure (e.g., the low-rank structure of approx. Hessian), we can address instances with millions of parameters. Our experiments demonstrate that \modelname~achieves superior accuracy compared to other pruning approaches within a fixed FLOP budget. For instance, for ResNet50 with 20\% of the total FLOPs retained, our approach improves the accuracy by 48\% relative to state-of-the-art. Furthermore, in gradual pruning settings with re-training between pruning steps, our framework outperforms existing pruning methods, emphasizing the significance of incorporating both FLOP and sparsity constraints for effective network pruning. 

\end{abstract}

\section{INTRODUCTION}

The remarkable success of modern neural networks has been accompanied by a surge in computational requirements ~\citep{devlin2018bert,brown2020language}, which pose significant challenges in deploying these models on resource-constrained devices such as mobile phones and Internet-of-things (IoT) devices. Network 
pruning is emerging as a promising framework towards mitigating
computational burdens. Pruning removes redundant or less important parameters, with the goal of retaining high performance while significantly reducing model size and computational complexity~\citep{blalock2020state}. 

Common approaches for network pruning include (i) impact-based \citep{lecun1989optimal,hassibi1992second,dong2017learning,singh2020woodfisher} and (ii) optimization-based methods \citep{yu2022combinatorial,benbaki2023fast}. Impact-based techniques eliminate weights based on the degree to which the removal of each individual weight might influence the loss function. Existing works mostly employ clever heuristics for evaluating impacts based on factors such as absolute values (also known as magnitude pruning, \citep{hanson1988comparing}), change in the loss function~\citep{lecun1989optimal,hassibi1992second}, or network connection sensitivity \citep{lee2018snip}. Despite their intuitive appeal, impact-based pruning may fall short of capturing the {\emph{joint}} effect of simultaneously removing multiple weights, resulting in suboptimal pruning outcomes. 
\cite{yu2022combinatorial} present optimization-based approaches for network pruning that consider the combined effect of pruning multiple weights at a time. Subsequently, CHITA~\citep{benbaki2023fast} formulate sparse pruning as an $\ell_0$-regularized sparse regression problem~\citep{hazimeh2020fast} and  
propose new optimization algorithms which are memory-efficient, and scalable, and result in state-of-the-art accuracy-sparsity tradeoffs on various examples. 
 

The approaches mentioned above aim to sparsify the network by reducing the number of non-zero (NNZ) weights, thus lowering the memory storage of the deployed model. However, these approaches do not directly consider other deployment costs or their proxies. For example, inference time and energy consumption are two important considerations arising in practical deployment scenarios, and the number of floating-point operations (FLOPs) has been proposed as a reasonable proxy for those two factors~\citep{yang2017designing}. 
While there is a sizable amount of work on network pruning to reduce NNZ, only a few studies attempt to directly reduce or control for the number of FLOPs during pruning. 
\citet{veniat2018learning,tang2018flops} propose dense-to-sparse training algorithms that incorporate FLOPs minimization by adding a weighted $\ell_0$-norm of weights to the optimization objective. They introduced stochastic gates to selectively mask weights, making the weighted $\ell_0$-regularized objective differentiable under certain distributions. \citet{singh2020woodfisher} design an impact-based pruning method using the empirical Fisher approximation of the Hessian matrix. They propose a FLOP-aware heuristic that 
considers the FLOPs for each parameter while pruning. 
While \citet{Kusupati2020STR} does not directly consider FLOPs as part of the objective, they achieve state-of-the-art accuracy-vs-FLOPs trade-offs by learning relatively uniform NNZ budgets across layers~\citep[Appendix S6]{singh2020woodfisher}.



In this paper, we introduce \modelname~(\underline{F}LOP-\underline{A}ware $\underline{\ell}_0$-based \underline{C}ombinatorial \underline{O}ptimization for \underline{N}etwork pruning), an efficient optimization-based framework for network pruning that considers both sparsity and FLOP constraints. Our optimization formulation allows us to directly adjust FLOP and NNZ budgets, resulting in pruned networks with a good accuracy-FLOPs-sparsity tradeoffs.
By effectively controlling for inference time-and-memory usage via their proxies NNZ-and-FLOPs, our approach can potentially allow practitioners to set and achieve their desired compression targets while retaining accuracy as much as possible. 

Magnitude pruning (MP) is a simple and popular method for pruning weights to reduce NNZ. We first generalize the notion of MP to accommodate NNZ and FLOP budgets simultaneously. We demonstrate that this generalized MP framework can be formulated as an integer linear program (ILP), and we develop an efficient algorithm with a linear convergence rate to obtain high-quality solutions for the problem. This provides insights into how the two constraints (NNZ and FLOP) interact with each other, thereby enhancing our understanding of the pruning problem. 


Obtaining a good solution to the ILP is an important component of \modelname. While parts of our framework (e.g., the local quadratic approximation based on Hessian)  \modelname~draw inspiration from recent studies~\citep{singh2020woodfisher, yu2022combinatorial,benbaki2023fast},
there are new contributions in this work. 
Current methods that address a sparsity constraint would not directly extend to handle the additional FLOP constraint. Therefore, to simultaneously address both FLOP and sparsity constraints, we explore new optimization techniques. 
In particular, we propose a discrete first-order (DFO) algorithm: in each iteration, we consider an ILP similar to the one arising in the generalized MP problem mentioned above. In addition, we leverage the low-rank structure of the approximated Hessian matrix for efficient optimization, bypassing the need 
for a costly Hessian computation and storage. 

Our experiments reveal that, given a fixed FLOP budget, our pruned models exhibit significantly better accuracy compared to other pruning approaches. We also conduct several ablation studies to highlight the importance of introducing joint budget constraints, as opposed to mere FLOP minimization. Moreover, when employed in a gradual pruning setting~\citep{Gale-state-sparsity,blalock2020state}, where re-training between pruning steps is performed, our pruning framework results in substantial performance gains compared to state-of-the-art pruning methods. 

\noindent \textbf{Contributions}~~ We summarize our contributions as follows:
\begin{itemize}[leftmargin=*,itemsep=1pt,topsep=0pt,parsep=0pt,partopsep=0pt]
   \item We introduce \modelname, a novel optimization-based framework for network pruning that accounts for both NNZ and FLOP budgets. Our approach achieves an optimal balance between NNZ and FLOP---proxies for inference time and memory usage---while preserving as much as possible the accuracy of highly compressed networks. 
   Our novel Discrete First-Order (DFO) algorithm iteratively 
   solves an ILP with both FLOP and NNZ constraints. By exploiting problem-structure, we can efficiently prune the network for large problem-sizes.  

\item We generalize the MP method to handle designated sparsity and FLOP constraints by formulating it as an ILP. We develop an efficient algorithm for solving the relaxed linear program (LP) and demonstrate that high-quality integer solutions can be recovered from LP solutions. This generalized MP framework is an important tool for FLOP-aware pruning, and  
plays a critical role in the DFO method of \modelname.

\item Our numerical results showcase \modelname's superior accuracy compared to other pruning approaches within a fixed FLOP budget. Notably, without retraining, \modelname~prunes a ResNet50 network to just 1.2 billion FLOPs (30\% of total FLOPs) with 73\% test accuracy, a mere 4\% reduction compared to the dense model, significantly outperforming state-of-the-art results (61\%). Moreover, in gradual pruning settings with re-training between pruning steps, our framework surpasses existing pruning methods. Our code is publicly available at:
\url{https://github.com/mazumder-lab/FALCON}.
\end{itemize}

%

This paper follows prior research on algorithms for unstructured network pruning. Our main focus is on proposing an optimization method that can handle the challenging task of unstructured pruning under both FLOP and NNZ constraints, useful proxies for inference time and memory requirements. Networks pruned via our novel framework can be deployed on specialized hardware, such as the efficient inference engine \citep{han2016eie}, to achieve noticeable enhancements in inference time. On a related note, a different line of work, known as structured pruning (see, e.g., \citet{he2017channel,guo2020dmcp}), seeks to remove entire components of the network, such as channels and neurons. This line of work is more readily suited to efficient practical hardware utilization but can potentially lead to greater accuracy loss for the same reduction in model size. The focus of our work here is on unstructured pruning.


\section{PROBLEM FORMULATION}\label{sect:prob}
We consider a neural network with a loss function defined as $\mathcal{L}(w)=(1/N)\sum_{i=1}^N\ell_i(w)$, where $w \in \R^p$ are the weights of the network, $N$ denotes the number of data points or samples, and $\ell_i(w)$ is a twice-differentiable function for the $i$-th sample.

\noindent\textbf{A weighted $\ell_0$ formulation of FLOPs}~~
The FLOPs of a neural network is the total number of floating point operations required for a single forward pass. We denote the FLOP cost for a parameter as the number of floating point operations associated with the parameter during a forward pass. For instance, the FLOPs of a scalar value within a kernel in a convolution layer can be calculated as $\text{FLOPs} = \text{(Output height)} \times \text{(Output width)}$. 

In our setting, we introduce the vector $f=(f_1,\dots,f_p)$, where the $i$-th element, $f_i$, represents the FLOP cost for the $i$-th parameter in the network.  We focus on the theoretical FLOP cost of the pruned network, which can be expressed as $\|w\|_{0,f}:=\sum_{i=1}^p f_i\textbf{1}_{w_i\neq 0}$. Here, $\textbf{1}_{w_i\neq 0}$ denotes the indicator function with a value of $1$ if $w_i$ is not pruned and $0$ otherwise. This formulation appears in earlier work such as \cite{singh2020woodfisher,Kusupati2020STR}.

\noindent\textbf{The structure of FLOP cost}~~
It is important to recognize that in most neural network architectures, including ResNet \citep{he2016deep} and MobileNet \citep{howard2017mobilenets} considered in our numerical experiments, all parameters within the same layer contribute equally to the total FLOP cost. We formalize this observation as the following fact:
\begin{fact}\label{ass:layer}
All parameters in the network can be divided into $L$ disjoint groups $C_1, C_2, \dots, C_L$ such that for any $j \in [L]$, all parameters within group $C_j$ has the same FLOP cost $f^j$. 
\end{fact}
We can divide the parameters into groups based on the layer they belong to, and the number of groups $L$ is usually much smaller than $p$ ($L\ll p$). Furthermore, we may improve this partition, as parameters across different layers may also share the same FLOP cost. This is an important observation that we will make use of both in our theoretical results and in our implementation.

\noindent\textbf{Pruning goal}~~Our goal is to minimize the number of non-zero parameters (NNZ) in the network and the FLOP cost during inference while preserving its performance as much as possible. Specifically, we are given a pre-trained weight vector $\bar w \in \R^p$, an NNZ budget $S$, and a FLOP budget $F$. We seek to compute a new weight vector $w \in \R^p$ that satisifes the following desiderata\footnote{We assume, without loss of generality, that $\{f_i\}_{i=1}^p$ are not identical. This differentiates the FLOP budget from the sparsity constraint.}:
\begin{itemize}[leftmargin=*,itemsep=1pt,topsep=0pt,parsep=0pt,partopsep=0pt]
\item \textbf{Retain model performance}: The loss function at $w$ should be as close as possible to the loss before pruning: $\mathcal{L}(w) \approx \mathcal{L}(\bar w)$.
\smallskip 
\item \textbf{FLOP budget}: The FLOP cost must be limited by the constraint $\|w\|_{0,f}=\sum_{i=1}^p f_i\textbf{1}_{w_i\neq 0} \leq F$.
\smallskip 
\item \textbf{Sparsity constraint}: The number of non-zero weights in $w$ must adhere to the NNZ budget: $\|w\|_0:=\sum_{i=1}^p \textbf{1}_{w_i\neq 0} \leq S$.
\end{itemize}

\noindent Throughout the paper, given a weight vector $w\in \R^p$, we denote its sparsity by $1-\|w\|_0/p$, and its NNZ by $\|w\|_0$. We refer to $S$ as the NNZ budget. We interchangeably use the terms sparsity constraint, cardinality constraint and budget constraint (on NNZ) to refer to the constraint $\|w\|_0\leq S$. 


\begin{figure}[!htbp]
\vspace{0mm}
   \begin{center}
    \includegraphics[width=0.7\columnwidth,trim=0 1cm 1cm 1cm]{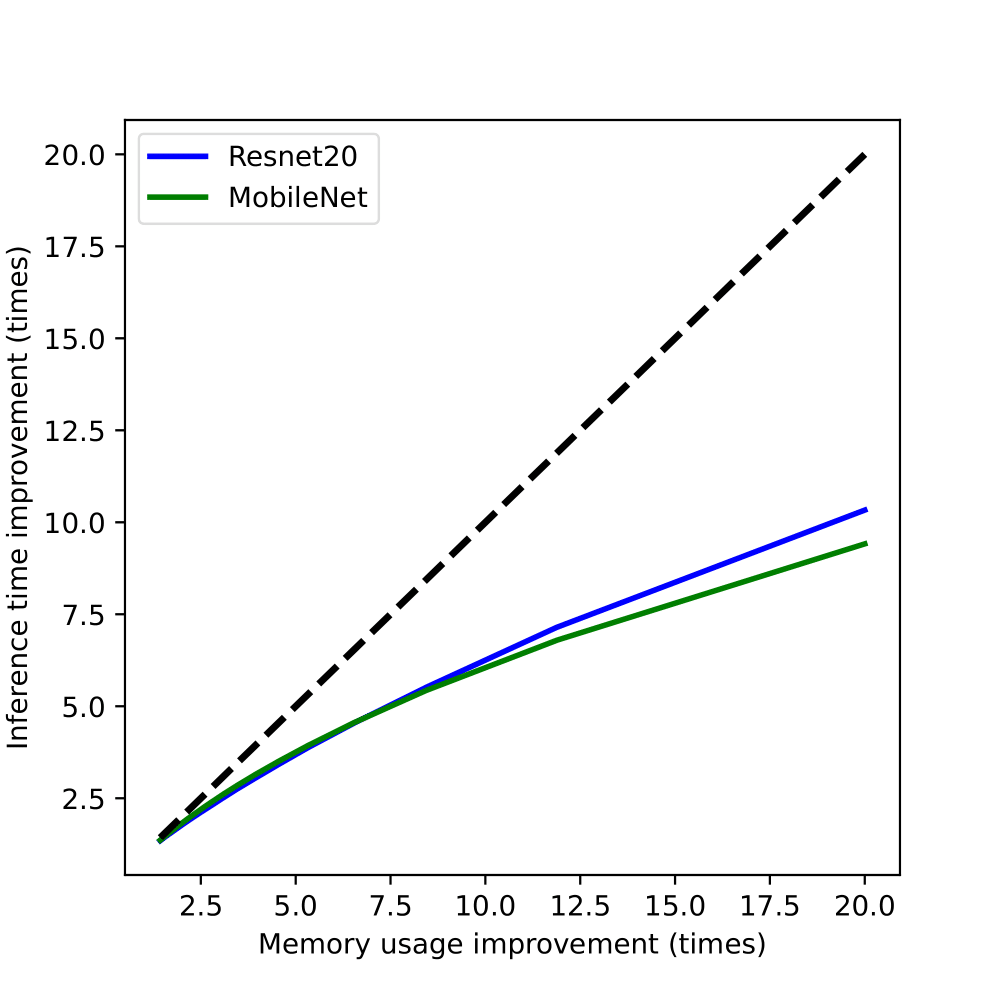}
  \end{center}
  \caption{Inference time improvement (measured by FLOPs) vs. memory usage improvement (measured by NNZ) of CHITA~\citep{benbaki2023fast}. Dashed black line denotes equal improvements. }
  \label{fig:motivation}
  \vspace{0mm}
\end{figure}
\textbf{A motivating example}~~ To motivate the use of the FLOP constraint, in Figure~\ref{fig:motivation}, we investigate the improvements of both inference time (measured by FLOPs) and memory usage (measured by NNZ) given by the pruned model. The pruning is done by a state-of-the-art sparse pruning method CHITA which aims to reduce the NNZ of the model. Figure~\ref{fig:motivation} shows that the model usually leads to much less improvement in terms of inference time (FLOPs) than in memory (NNZ). In particular, although a highly pruned model can lead to a 20$\times$ improvement in memory, its actual inference time improvement is only 8-10$\times$. This suggests that NNZ alone is not a good network pruning metric for inference time improvement---this motivates us to incorporate the FLOP constraint to accompany the commonly-used sparsity constraint in the existing literature. Furthermore, as evidenced in Section~\ref{sect:exp-joint}, combining sparsity and FLOP constraints allows for a relatively uniform sparsity across layers of the network, leading to improved accuracy. 

\section{ILP-BASED MAGNITUDE PRUNING UNDER NNZ AND FLOP BUDGETS }\label{sect:impact}

Magnitude pruning, a popular method in the neural network pruning literature, takes a simple form when addressing the sparsity constraint alone. In particular, in the context of unstructured sparsity, this eliminates a portion of parameters with the smallest absolute values while retaining the remaining weights. However, the pruning process becomes more complex when we consider both FLOP and NNZ constraints. To see this, we will first consider a magnitude-based pruning framework incorporating both constraints --- this
sheds insights into our proposed optimization formulation of network pruning with both FLOP and sparsity constraints (cf Section \ref{sect:iht}).

\subsection{Integer Linear Programming (ILP) Formulation}
The MP approach for sparse pruning is to select a set of parameters with the least absolute values to remove while adhering to cardinality constraints. For each $i\in[p]$, let the binary variable $z_i = \textbf{1}_{w_i \neq 0}$ denote if weight $w_i$ is retained or pruned, and $I_i=\bar w_i^2$ denote the squared magnitude of the $i$-th parameter in the network. Then, the MP method is equivalent to solving the following cardinality-constrained optimization problem:
\begin{equation}\label{eq:MP}
\max_{z\in\{0,1\}^p}~~ Q_I(z):=\textstyle\sum\limits_{i=1}^p I_i z_i, ~~
\text{s.t.}~~\sum\limits_{i=1}^p z_i \le S,
\end{equation}
which can be readily solved via sorting (see Appendix \ref{sect:auxthm}):
After Problem~\eqref{eq:MP} is solved for $z_i$'s, the pruned weights obtained by MP are given as $w_i = \bar w_iz_i$ for all $i$.

Here, we generalize the MP approach in Problem~\eqref{eq:MP} to include an additional FLOP constraint in the model to 
cater to both FLOP and NNZ budget constraints. 
Note that the number of theoretical FLOPs during inference can be expressed as $\sum_{i=1}^p f_iz_i$. This leads to the following ILP problem:
\begin{equation}\label{eq:main-impact}\tag{ILP}
\max_{z\in\{0,1\}^p}~~\textstyle\sum\limits_{i=1}^p I_i z_i, 
~~\text{s.t.}~~  \sum\limits_{i=1}^p f_iz_i \le F,~~ \sum\limits_{i=1}^p z_i \le S.    
\end{equation}
When the FLOP budget $F$ is very large, the FLOP constraint in \eqref{eq:main-impact} becomes redundant, i.e., including the FLOP constraint does not affect the pruning decision. In this case, the problem reduces to magnitude pruning with only a sparsity constraint. On the other hand, if the sparsity constraint is redundant, \eqref{eq:main-impact} simplifies to magnitude pruning focused exclusively on FLOP cost. The \eqref{eq:main-impact} model allows for fine-grained control over FLOP and NNZ when both constraints are present. This scenario achieves the highest accuracy (under a fixed FLOP budget), as demonstrated in Section~\ref{sect:exp-joint}. 


\subsection{Solving the Relaxed Problem}
Despite its seemingly simple structure, solving \eqref{eq:main-impact} is challenging due to its discrete nature. When both FLOP and sparsity constraints are considered, pruning becomes more challenging compared to Problem~\eqref{eq:MP}, as there is no closed-form optimal solution. While integer programming can, in principle, be solved for small-to-moderate size problems using powerful commercial solvers (e.g., Gurobi), we do not pursue this approach since the number of parameters $p$ in the network  can reach millions in practice, posing significant computational challenges.

For computational reasons, we consider a convex relaxation of~\eqref{eq:main-impact}, resulting in the following Linear Program (LP):
\begin{equation}\label{eq:lp-impact}\tag{RLP}
\max_{z\in[0,1]^p}~~ \textstyle\sum\limits_{i=1}^p I_i z_i, ~~
\text{s.t.} ~~  \sum\limits_{i=1}^p f_iz_i \le F,~~ \sum\limits_{i=1}^p z_i \le S.
\end{equation}
Problem \eqref{eq:lp-impact} relaxes binary variables $\{z_i\}_{i=1}^p$ in~\eqref{eq:main-impact} to continuous variables in $[0,1]$. We will discuss in Theorem~\ref{thm:optgap} how to retrieve a feasible binary solution from the relaxed problem and analyze the gap between that and the optimal solution to \eqref{eq:main-impact}.

However, due to its large scale, solving \eqref{eq:lp-impact} using commercial or open-source LP solvers is still time-consuming. To address this issue, we design an efficient custom solver for \eqref{eq:lp-impact} by leveraging its dual problem, given by:
 \begin{equation}\label{eq:lp-dual}
    \begin{aligned}
        \min_{\lambda_1 \geq 0,\lambda_2 \geq 0} \,\, & D(\lambda_1,\lambda_2):=S\lambda_1 + F\lambda_2 \\
        &\qquad\qquad\,\,\, + \textstyle\sum\limits_{i=1}^p \max\{I_i-\lambda_1-f_i\lambda_2,0\}. 
    \end{aligned}
\end{equation}
A key observation is that for a fixed $\lambda_2$, minimization over $\lambda_1$ can be computed exactly as 
\begin{equation}\label{eq:optlambda1}
    \argmin_{\lambda_1\ge 0}D(\lambda_1,\lambda_2) = \max\{ (I-\lambda_2f)_{(S)},0\},
\end{equation}
where $I=(I_1,\dots,I_p)$, and $(v)_{(S)}$ denote the $S$-th largest element of a vector $v$. Importantly, the resulting function $g(\lambda_2):=\min_{\lambda_1\ge 0}D(\lambda_1,\lambda_2)$ is a one-dimensional convex function (refer to \cite[Chapter 3]{boydvandenberghe2004} for a proof). As a consequence, the dual problem \eqref{eq:lp-dual} can be efficiently solved using a golden-section search method \cite[Chapter 10]{yao2007early}. The details of this procedure are outlined in Algorithm \ref{alg:proj}.

\noindent\textbf{Optimized sorting method}~~
The main computational load in each iteration of Algorithm \ref{alg:proj} stems from two steps: identifying the $S$-th largest element in the vector $(I-\lambda_2f)$ and calculating $g(\lambda_2)$ according to \eqref{eq:optlambda1}. Using conventional methods (e.g., quicksort) for calculating the $S$-th largest element demands $O(p)$ time. Leveraging Fact 2.1, we can partition $(I-\lambda_2f)$ into $L$ sorted arrays for any $\lambda_2$, with no added cost except initial preprocessing. We introduce a novel algorithm that leverages this structure to reduce the time complexity of finding $(I-\lambda_2f)_{(S)}$ and $g(\lambda_2)$ to $O\big(L(\log p)^2\big)$. In mny practical scenarios, $L$ often remains under $100$---in such cases we observe useful speedups by a factor of dozens through our novel approach. Readers can refer to Appendix \ref{subsec:prooftime} for a detailed description of our proposed approach.

\begin{algorithm}[h]
\begin{algorithmic}[1]
\REQUIRE NNZ budget $S$, FLOP budget $F$, magnitude $I_i$ and FLOP cost $f_i$ of each parameter, and accuracy $\varepsilon$.
\STATE Initialize $\lambda_2^{min}= 0,\,\,\lambda_2^{max}=\max_{i\in [p]}\{ I_i/f_i\}$ and $\alpha=\frac{3-\sqrt 5}{2}$.
\STATE  Set $
        \lambda_2 = \lambda_2^{min} + \alpha (\lambda_2^{max}-\lambda_2^{min}),\quad
        \lambda_2'= \lambda_2^{max} - \alpha (\lambda_2^{max}-\lambda_2^{min}).$
\WHILE {$\lambda_2^{max}-\lambda_2^{min} > \varepsilon$}
\STATE Compute $g(\lambda_2)$ and $g(\lambda_2')$ via efficient method discussed in Appendix \ref{subsec:prooftime}.
\IF {$g(\lambda_2) \le g(\lambda_2')$}
\STATE Update $\lambda_2^{max}= \lambda_2'$,\,\,\,$\lambda_2'= \lambda_2$, and $\lambda_2= \lambda_2^{min} + \alpha (\lambda_2^{max}-\lambda_2^{min})$.
\ELSE
\STATE Update $\lambda_2^{min}= \lambda_2$,\,\,\,$\lambda_2= \lambda_2'$, and $\lambda_2'=  \lambda_2^{max} - \alpha (\lambda_2^{max}-\lambda_2^{min})$.
\ENDIF
\ENDWHILE
\end{algorithmic}
\caption{Solving Problem \eqref{eq:lp-dual} via golden-section search}
\label{alg:proj}

\end{algorithm}

\subsection{Theoretical Results} \label{subsec:theorem}
The following theorem presents the cost of solving Problem~\eqref{eq:lp-dual} by Algorithm \ref{alg:proj}.
\begin{theorem}\label{thm:complexity}
Making use of Fact \ref{ass:layer}, each iteration of Algorithm \ref{alg:proj} takes $O\big(L(\log p)^2\big)$ time complexity. Moreover, for $\epsilon>0$, it takes $O\big(p\log p+ L(\log p)^2\log(1/\varepsilon)\big)$ time to compute a $\varepsilon$-accurate solution of the dual problem \eqref{eq:lp-dual}.
\end{theorem}

Since Algorithm \ref{alg:proj} achieves a linear convergence rate, we can reasonably assume that obtaining the optimal solution for the dual problem requires minimal cost in practice. Theorem \ref{thm:optgap} ensures that, given an optimal solution to the dual problem, we can recover a high-quality feasible solution for the original integer program~\eqref{eq:main-impact}. The proofs for Theorem \ref{thm:complexity} and Theorem \ref{thm:optgap} can be found in Appendix~\ref{app:proofs}.

\begin{theorem}\label{thm:optgap}
We assume Fact~\ref{ass:layer} holds true, and denote $L_f = \sum_{j=1}^L f^j$. Given an optimal solution of the dual problem \eqref{eq:lp-dual}, we can compute a feasible binary solution $\hat z$ to \eqref{eq:main-impact} with the following optimality gap:
\begin{equation}
[Q_I^*-Q_I(\hat z)]/Q_I^* \le \max\{ L/S,L_f/F\},
\end{equation}
where $Q_I^*$ is the optimal objective of \eqref{eq:main-impact}.
\end{theorem}
In most network architectures, the value of $S$ is often in the millions, while the number of layers (which serves as the upper bound of $L$) is limited to a few hundred. Hence, $L/S$ is approximately $10^{-4}$. Similarly, the value of $L_f/F$ is typically of the order of $10^{-4}$. This suggests that the relaxed problem approximates Problem~\eqref{eq:main-impact} with a medium-to-high accuracy.

We note that the algorithm and theorems presented in this section apply to any choice of $\{I_i\}_{i=1}^p$ in \eqref{eq:main-impact} as long as $I_i\ge 0$. Hence, our framework can be potentially applied to impact-based pruning with FLOP and NNZ constraints by defining $I_i$ as the impact of the $i$-th parameter.


\section{OPTIMIZATION FORMULATION FOR PRUNING}\label{sect:iht}
In this section, we present our algorithmic framework \modelname~ for pruning a neural network given constraints on the model's sparsity (NNZ) and FLOPs. In Section~\ref{subsect:ihtobj}, we formulate the pruning problem as a sparse regression problem with both $\ell_0$ and weighted $\ell_0$ constraints. In Section~\ref{subsect:ihtmethod}, we propose a discrete first-order method (DFO) to handle both sparsity and FLOP constraints. This DFO method requires repeatedly solving Problem~\eqref{eq:main-impact} proposed in the last section. 

\subsection{Optimization Formulation with NNZ and FLOP Constraints}\label{subsect:ihtobj}

Our optimization-based framework builds upon earlier work~\citep{lecun1989optimal,hassibi1992second,singh2020woodfisher} that use a local model $\mathcal{L}$ around the pre-trained weights $\bar w$:
\begin{equation}
\label{eq:local-quadratic}
\begin{aligned}
    \mathcal{L}(w)&=\mathcal{L}(\bar w)+\nabla \mathcal{L}(\bar w)^\top (w-\bar w) \\
    &+ \frac12(w-\bar w)^\top \nabla^2\mathcal{L}(\bar w)(w-\bar w)+O(\|w-\bar w\|^3).
\end{aligned}
\end{equation}


By selecting appropriate gradient and Hessian approximations $g\approx\nabla\mathcal{L}(\bar w), H\approx\nabla^2\mathcal{L}(\bar w)$ and disregarding higher-order terms, we derive $Q_{L_0}(w)$ as a local approximation of the loss $\mathcal{L}$:
\begin{equation}\label{eq:local-quadratic2}
Q_{L_0}(w) :=\mathcal{L}(\bar w)+ g^\top(w-\bar w)+\frac12(w-\bar w)^\top H(w-\bar w).
\end{equation} 
Following recent works~\citep{singh2020woodfisher,benbaki2023fast}, we approximate the gradient using the stochastic gradient on $n$ samples: 
\begin{equation}\label{eq:grad}
g=(1/n)\sum_{i=1}^n\nabla \ell_i(\bar w)=\displaystyle (1/n) X^\top e\in\R^p.
\end{equation}
We approximate the Hessian matrix using the empirical Fisher information derived from the same $n$ samples, i.e. we take:
\begin{equation}\label{eq:Hessian}
H=\frac1n\textstyle\sum\limits_{i=1}^n\nabla \ell_i(\bar w)\nabla \ell_i(\bar w)^\top=\displaystyle\frac1n X^\top X\in\R^{p\times p},
\end{equation}
where $X=[\nabla \ell_1(\bar w),\ldots, \nabla \ell_n(\bar w)]^\top\in\R^{n\times p}$. 
It's important to observe that matrix $H$ has a rank of $n$. In practice, $p$ could reach millions while $n$ remains below thousands. Leveraging the low-rank property of $H$ offers significant reductions in memory usage and runtime, as elaborated in the following discussion.

Putting together the pieces, our optimization model is formulated as follows:
\begin{equation}\label{eq:miqp-main}
\begin{aligned}
\min_w ~~~ &Q_{L}(w):= g^\top(w-\bar w)+\frac12(w-\bar w)^\top H(w-\bar w) \\ &\qquad\quad\, + \frac{n\lambda}{2}\|w-\bar w\|^2,\\
\text{s.t.}~~~&\|w\|_0\le S,\,\,\,\,\|w\|_{0,f}\le F,
\end{aligned}
\end{equation}
Here, $\lambda\geq 0$ represents the strength of the ridge regularization, while $\|w\|_0\le S$ and $\|w\|_{0,f}\le F$ are two combinatorial constraints related to NNZ and FLOP budgets, respectively. 


By leveraging the low-rank structure of the Hessian matrix, we can express our proposed  formulation  \eqref{eq:miqp-main} in a Hessian-free form. This eliminates the need to store the expensive full dense $p \times p$ Hessian matrix---instead, we only need to store  a much smaller $n \times p$ matrix $X$. 
This results in a substantial reduction in memory usage.  In contrast, another FLOPs-aware pruning method \citep{singh2020woodfisher} employs a dense $p \times p$ matrix as an approximation of the Hessian, which can be prohibitively expensive in terms of both runtime and memory. For completeness, all details on the formulation are provided in Appendix~\ref{subapp:formulation-details}.


Problem \eqref{eq:miqp-main} can be formulated as a mixed integer quadratic program which is challenging to solve. In contrast to previous studies that consider a single cardinality constraint, here we are dealing with an additional FLOP constraint. This necessitates developing new algorithms, as we discuss below.


\subsection{A Modified DFO Method for Solving \eqref{eq:miqp-main}} \label{subsect:ihtmethod}

We outline the primary concepts behind our suggested algorithm for problem~\eqref{eq:miqp-main} and provide further details in Appendix~\ref{subapp:algorithm-details}.

Our optimization approach is based on the Discrete First-order (DFO) method~\citep{bertsimas2016best}, which optimizes~\eqref{eq:miqp-main} using an iterative process. Every iteration, referred to as a \emph{DFO update}, concurrently updates the support and weights. We discuss below how this DFO update is equivalent to solving the integer program~\eqref{eq:main-impact}.

Taking advantage of the low-rank structure, we can bypass the need to compute the entire Hessian matrix, resulting in important cost savings. While the basic version of the DFO algorithm may be slow for problems involving millions of parameters, we enhance our algorithm's computational performance by incorporating an active set strategy and schemes to update the weights on nonzero weights after support stabilization. These enhancements substantially improve computational efficiency and solution quality, making our method suitable for large-scale network pruning problems.

\noindent\paragraph{DFO update}~~
The DFO algorithm operates by taking a gradient step with step-size $\tau^s$ from the current iteration and projecting it onto the feasible set (see update~\eqref{eq:iht} below). For any vector $\bar x$, the projection operator $P_{S,F}(\bar x)$ is defined as
\begin{equation}\label{eq:dfoproj}
    P_{S,F}(\bar x) = \argmin_x~ \|x-\bar x\|^2 ~~\text{s.t.}~~~ \|x\|_0\le S,~~ \|x\|_{0,f}\le F.
\end{equation}


$P_{S,F}(\bar x)$ computes the feasible point to problem \eqref{eq:miqp-main} that is closest to $\bar x$. Crucially, the projection can be efficiently solved using the ILP discussed in Section \ref{sect:impact}, as stated in the following lemma: 
\begin{lemma}\label{lemma:proj}
    Solving the projection problem \eqref{eq:dfoproj} is equivalent to solving \eqref{eq:main-impact} with $I_i=(\bar x_i)^2$ for $i\in [p]$. Given a solution $z^*$ of \eqref{eq:main-impact}, 
\begin{equation}
x_i=\bar x_i z_i^*,~\forall i\in[p]
\end{equation}
provides a solution to the projection problem \eqref{eq:dfoproj}. 
\end{lemma}
The proof of the Lemma is provided in Appendix \ref{sect:auxthm}. By our efficient approach (i.e., Algorithm \ref{alg:proj}) for \eqref{eq:main-impact}, we can solve the projection problem to near-optimality quickly.

Applying DFO to problem \eqref{eq:miqp-main} results in the following update:
\begin{equation}\label{eq:iht}
    \begin{aligned}
         w^{t+1} &= \textsf{DFO}(w^t,\tau^s) := P_{S,F}\left(w^t-\tau^s \nabla Q_L(w^t)\right)\\
    \end{aligned}
\end{equation}
where $\tau^s>0$ is a suitable stepsize. Due to the low-rank structure of $H$, matrix-vector multiplications with $H$ in $\nabla Q_L(w^t)$ can be performed by operating on matrix $X\in \mathbb{R}^{n\times p}$ with cost $O(np)$---see Appendix~\ref{subapp:algorithm-details} for details.

\noindent\textbf{Convergence result}~~
Our DFO method can be viewed as an inexact proximal gradient method. At each iteration, we convert the projection operation in \eqref{eq:iht} into \eqref{eq:main-impact} (this is an exact reformulation) and solve \eqref{eq:main-impact} approximately via Algorithm \ref{alg:proj}. As per \citet{gu2018inexact}, for convergence, we need the error in the approximate solution to \eqref{eq:main-impact} to be summable (across the iterations $t$). This can be achieved for example, by solving \eqref{eq:main-impact} to a higher accuracy (as $t$ increases). The latter can be done with off-the-shelf IP solvers with which we can control the gap from the optimal objective value. In practice, we would need an IP solver to refine (or improve) the rounding-based solution derived from Algorithm \ref{alg:proj}. In this paper we do not further investigate the (theoretical) convergence guarantees of our algorithm because our experimental results appear to suggest that applying \eqref{eq:iht} (with the projection operator solved approximately by Algorithm \ref{alg:proj}) for a fixed number of iterations already yields solutions of high quality. For more detailed convergence properties of inexact proximal gradient methods, please refer to \citet{schmidt2011convergence,gu2018inexact}.

\noindent\textbf{Enhancing DFO efficiency}~~
To improve the efficiency of the proposed DFO method, we adopt an active set strategy \citep{nocedal1999numerical,hazimeh2020fast} that limits DFO updates to a small subset of variables on active set---this brings down the cost of every iteration from $O(pn)$ to approximately $O(Sn)$. We further accelerate the convergence of the DFO method by using a back-solve procedure to refine the nonzero coefficients. Detailed discussions on active set strategy and back-solve procedure are available in Appendix~\ref{subsec:active-set}.

\noindent\textbf{A multi-stage procedure}~~
The DFO method introduced above delivers high-quality feasible solutions to optimization problem \eqref{eq:miqp-main}. We refer to this as a single-stage approach, as it optimizes the local quadratic approximation to the loss function $\mathcal L$ for \emph{one} time. However, the performance of the resulting pruned network is sensitive to the quality of the local quadratic approximation, and we observe a degrading accuracy when we aim for high sparsity and low FLOPs.  To mitigate this, we propose a multi-stage process called \modelname++ that operates on a better approximation of the loss $\mathcal L$. Here we iteratively refine the local quadratic models and solve them with gradually decreasing NNZ and FLOP budgets, thereby preserving efficiency while improving accuracy. The multistage approach extends the framework of \cite{singh2020woodfisher} to incorporate both FLOP and NNZ constraints. Different from gradual pruning~\citep{han2015learning},  \modelname++ avoids the need for computationally intensive fine-tuning via SGD. 
We present the details of \modelname++ in Algorithm \ref{alg:falcon}. Note that  \modelname~(i.e., the single stage approach) can be viewed as a special case of \modelname++ by setting the number of stages $T_0=1$ in Algorithm \ref{alg:falcon}. More details of our multi-stage approach are provided in Appendix~\ref{subsec:multistage}. 

\begin{algorithm}[!htb]
\begin{algorithmic}[1]
\REQUIRE The pre-trained weights $\bar w$, target NNZ budget $S$ and FLOPs budget $F$, and the number of stages $T_0$.
\STATE Set $w^0=\bar w$; construct sequences of parameters with decreasing NNZ and FLOPs budgets as follows:
{\small
\begin{equation*}
    S_1 \ge S_2\ge\cdots\ge S_{T_0}=S;\,\,\,F_1 \ge F_2\ge\cdots\ge F_{T_0}=F
\end{equation*} }
\FOR {$t=1,2,\ldots,T_0$}
\STATE At current solution $w^{t-1}$, calculate the stochastic gradient on a batch of $n$ training points 
\STATE Construct the objective $Q_L(w)$ in \eqref{eq:miqp-main} using Hessian approximation \eqref{eq:Hessian} and gradient approximation \eqref{eq:grad}.
\STATE Obtain a solution $w^t$ to problem~\eqref{eq:miqp-main} with sparsity budget $S_t$ and FLOPs budget $F_t$ by performing DFO updates \eqref{eq:iht} with the active set strategy and back-solve procedure (see Appendix \ref{subapp:algorithm-details} for details).
\ENDFOR
\end{algorithmic}
\caption{\modelname++: a multi-stage procedure for pruning networks under both NNZ and FLOPs budgets}
\label{alg:falcon}
\end{algorithm}

\section{NUMERICAL EXPERIMENTS}\label{sec:expts}
In this section, we compare our proposed algorithms with existing start-of-the-art approaches in two scenarios:~(i)~ one-shot pruning in which the model is pruned only once after it has been fully trained, and~(ii)~gradual pruning, also known as re-training, which iteratively prunes and fine-tunes the model over a period of time. We evaluate our proposed framework \modelname~on various pre-trained networks including ResNet20~(\citealp{he2016deep}, 260k parameters) trained on CIFAR10~\citep{krizhevsky2009learning}, MobileNet (\citep{howard2017mobilenets}, 4.2M parameters) and ResNet50 (25.6M parameters) trained on ImageNet~\citep{deng2009imagenet}. 
Detailed information on the experimental setup and reproducibility can be found in Appendix~\ref{subsec:expt-setup}. Ablation studies and more experimental results are shown in Appendix \ref{subsec:addexp}.

\subsection{One-shot Pruning}\label{subsec:oneshot}

\subsubsection{Accuracy given the FLOP budget}
To assess the effectiveness of our proposed frameworks \modelname~ and \modelname++, we compare them with several leading one-shot pruning methods. These include  MP~\citep{mozer1989using}, WF (with FLOP-aware pruning statistic)~\citep[Appendix S6]{singh2020woodfisher}, CHITA~\citep{benbaki2023fast}. All these approaches have been used in sparse pruning, though none of these methods consider directly minimizing both FLOPs and NNZs. For a fair comparison, the NNZ budget $S$ for these methods was calibrated such that the pruned networks meet the FLOP budget.

\begin{table}[tbp]
    \centering
\caption{The pruning performance (accuracy) of various methods on ResNet20, MobileNetV1, and ResNet50. The bracketed number in the FLOPs column indicates the proportion of FLOPs needed for inference in the pruned network versus the dense network. We take five runs for our single-stage (\modelname) and multi-stage (\modelname++) approaches and report the mean and standard error~(in the brackets). The {\color{red}\textbf{first}} and {\color{blue}\textbf{second}} best accuracy values are highlighted in bold.  } 
\label{tab:accuracy}
    \resizebox{0.95\columnwidth}{!}
    {\begin{tabular}{c|c|ccc|cc}
    \toprule
\footnotesize Network & FLOPs & MP  & WF & CHITA & \modelname & \modelname++ \\
\midrule
\multirow{6}{*}{
\begin{minipage}{2cm}
\begin{center}
    ResNet20\\
    on CIFAR10\\
    (91.36\%)
\end{center}
\end{minipage}
}& 24.3M\,(60\%) &88.89 &91.13 &90.95 &{\color{blue} \textbf{91.38(±0.10)}} &{\color{red} \textbf{91.39(±0.10)}}\\ 
& 20.3M\,(50\%) &85.95 &90.42 &89.86 &{\color{blue} \textbf{90.87(±0.09)}} &{\color{red} \textbf{91.07(±0.13)}}\\ 
& 16.2M\,(40\%) &80.42 &87.83 &86.10 &{\color{blue} \textbf{89.67(±0.18)}} &{\color{red} \textbf{90.58(±0.17)}}\\ 
& 12.2M\,(30\%) &58.78 &77.43 &72.48 &{\color{blue} \textbf{84.42(±0.70)}} &{\color{red} \textbf{89.64(±0.26)}}\\ 
& 8.1M\,(20\%) &15.04 &39.21 &30.07 &{\color{blue} \textbf{65.17(±3.95)}} &{\color{red} \textbf{87.59(±0.16)}}\\ 
& 4.1M\,(10\%) &10.27 &12.31 &11.61 &{\color{blue} \textbf{19.14(±2.25)}} &{\color{red} \textbf{81.60(±0.37)}}\\ 
\midrule
\multirow{6}{*}{
\begin{minipage}{2cm}
\begin{center}
    MobileNetV1\\
    on ImageNet\\
    (71.95\%)
\end{center}
\end{minipage}
}& 398M\,(70\%) &70.89 &71.70 &71.82 &{\color{blue} \textbf{71.83(±0.05)}} &{\color{red} \textbf{71.91(±0.01)}}\\ 
& 341M\,(60\%) &67.34 &70.95 &71.25 &{\color{blue} \textbf{71.42(±0.04)}} &{\color{red} \textbf{71.50(±0.04)}}\\ 
& 284M\,(50\%) &51.13 &68.32 &68.78 &{\color{blue} \textbf{70.35(±0.10)}} &{\color{red} \textbf{70.66(±0.06)}}\\ 
& 227M\,(40\%) &14.85 &55.76 &65.03 &{\color{blue} \textbf{67.18(±0.24)}} &{\color{red} \textbf{68.89(±0.03)}}\\ 
& 170M\,(30\%) &0.65 &18.72 &50.89 &{\color{blue} \textbf{58.40(±0.31)}} &{\color{red} \textbf{64.74(±0.13)}}\\ 
& 113M\,(20\%) &0.10 &0.19 &3.92 &{\color{blue} \textbf{25.82(±2.09)}} &{\color{red} \textbf{53.84(±0.11)}}\\ 
\midrule
\multirow{6}{*}{
\begin{minipage}{2cm}
\begin{center}
    ResNet50\\
    on ImageNet\\
    (77.01\%)
\end{center}
\end{minipage}
}& 2.3G\,(60\%) &75.38 &76.67 &76.56 &{\color{blue} \textbf{76.86(±0.02)}} &{\color{red} \textbf{76.89(±0.03)}}\\ 
& 2.0G\,(50\%) &70.85 &75.76 &75.22 &{\color{blue} \textbf{76.39(±0.02)}} &{\color{red} \textbf{76.46(±0.07)}}\\ 
& 1.6G\,(40\%) &62.37 &72.65 &73.32 &{\color{blue} \textbf{75.28(±0.05)}} &{\color{red} \textbf{75.64(±0.06)}}\\ 
& 1.2G\,(30\%) &22.43 &58.78 &64.05 &{\color{blue} \textbf{71.54(±0.15)}} &{\color{red} \textbf{73.49(±0.15)}}\\ 
& 817M\,(20\%) &0.56 &5.39 &12.88 &{\color{blue} \textbf{54.27(±0.44)}} &{\color{red} \textbf{67.08(±0.20)}}\\ 
& 408M\,(10\%) &0.10 &0.10 &0.10 &{\color{blue} \textbf{0.46(±0.07)}} &{\color{red} \textbf{47.63(±0.38)}}\\ 
\bottomrule
 \end{tabular} }
\end{table}

Table~\ref{tab:accuracy} compares the test accuracy for pruned ResNet20, MobileNetV1, and ResNet50 under varying FLOP budgets. Our \modelname~ framework delivers considerably superior accuracy relative to existing methods. Furthermore, our multi-stage method: \modelname++, surpasses other techniques substantially, without incurring the extra cost of re-training.

\subsubsection{Quantifying the usefulness of having both FLOP and NNZ budget constraints}\label{sect:exp-joint}
Our framework \modelname, can handle both sparsity and FLOP constraints, thereby enabling practitioners to effectively set and achieve their model compression objectives while retaining model accuracy as much as possible. In principle, 
as we take into account both FLOPs and sparsity (NNZ) constraints, we expect \modelname~ to result in a model outperforming those pruned solely under FLOPs or sparsity constraints.

\begin{figure}[tbp]
\vspace{0mm}
   \centering
   \begin{minipage}{0.48\linewidth}
        \includegraphics[width=0.95\columnwidth,trim=0.5cm 0.5cm 0.5cm 0.5cm]{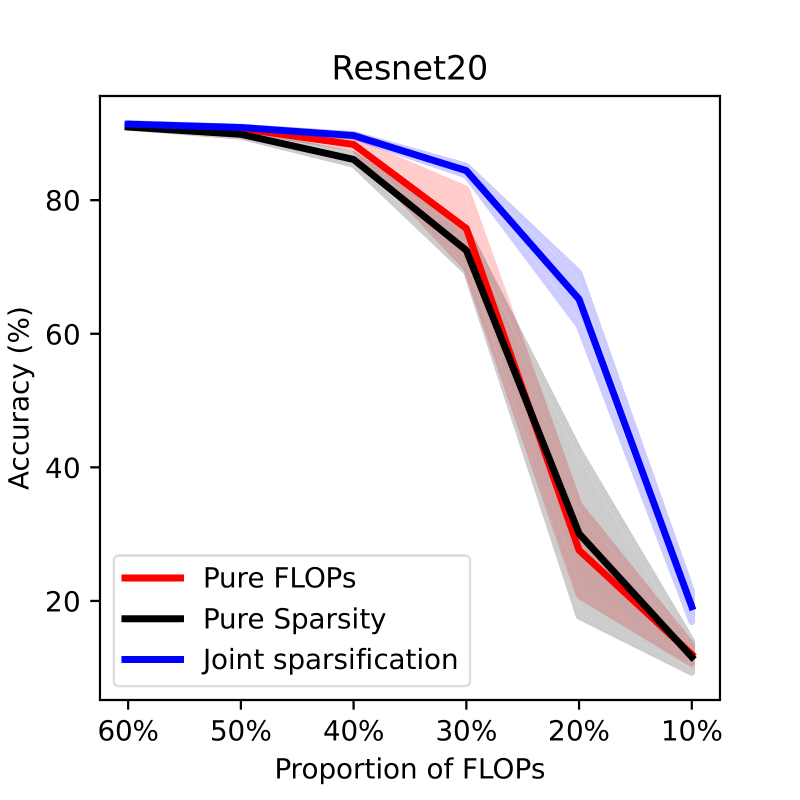}
    \end{minipage}%
    \begin{minipage}{0.48\linewidth}
        \includegraphics[width=0.95\columnwidth,trim=0.5cm 0.5cm 0.5cm 0.5cm]{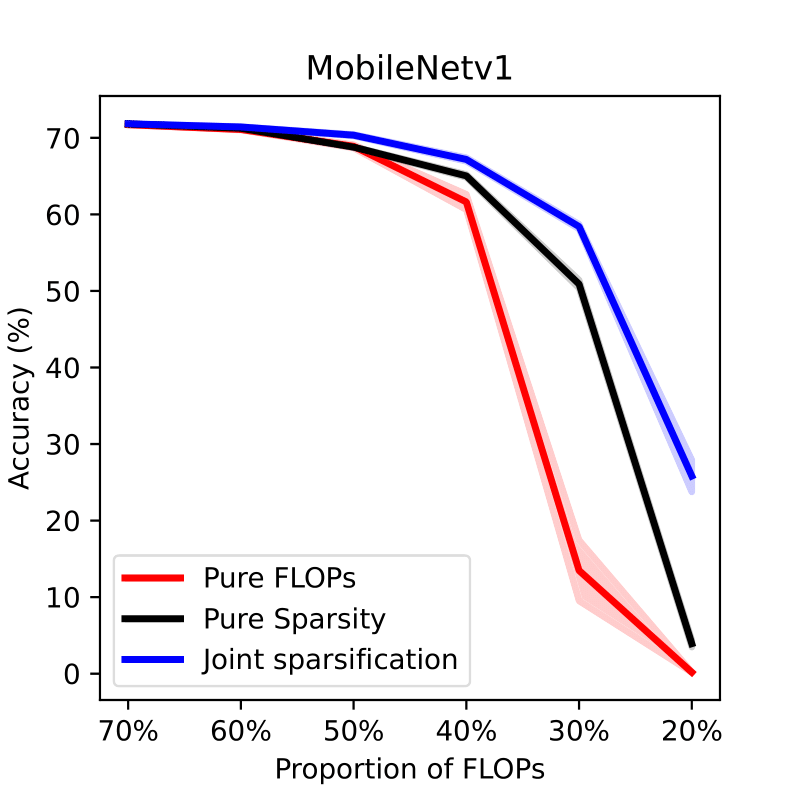}
    \end{minipage}%
    \hfill
    \begin{minipage}{0.48\linewidth}
        \includegraphics[width=0.95\columnwidth,trim=0.5cm 0.5cm 0.5cm 0.5cm]{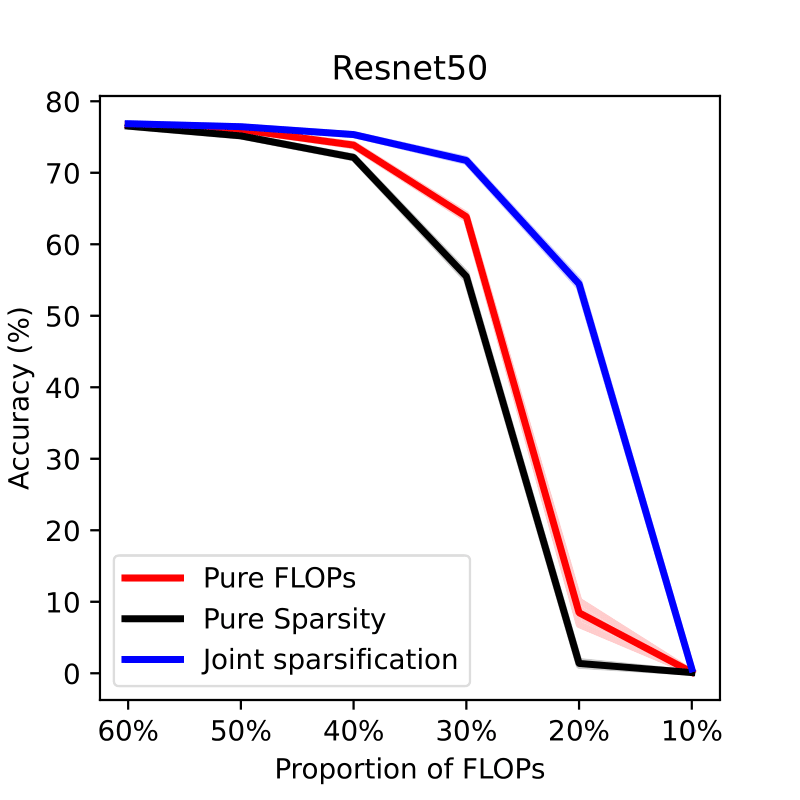}
    \end{minipage}%
    \begin{minipage}[c]{0.48\linewidth}
    \vspace{5mm}
       \caption{ Accuracy of the network pruned by \modelname~across three scenarios: pure FLOP constraint, pure sparsity constraint, and joint sparsification. The error bar represents the standard error over five runs.}
       \label{fig:joint}
    \end{minipage}%
    \hfill
    
\end{figure}

To illustrate this, we evaluate \modelname~under a fixed FLOP budget ($F_0$) across three distinct scenarios.
(i)~Pure FLOP constraint: we set the FLOP budget to $F=F_0$ and the NNZ budget to $S=\infty$; (ii) Pure sparsity constraint: we fix $F=\infty$ and find the NNZ budget $S$ so that the FLOPs of the resulting network precisely equals $F_0$; (iii)~Joint sparsification: we choose $F=F_0$ and $S$ optimally to maximize accuracy. As depicted in Figure \ref{fig:joint}, joint sparsification notably improves accuracy under a fixed FLOP budget compared to models pruned only based on a FLOP or NNZ constraint. 

\begin{figure}[H]
\vspace{0mm}
   \centering
   \begin{minipage}{0.48\linewidth}
        \includegraphics[width=0.95\columnwidth,trim=0.3cm 0.5cm 0.8cm 0.5cm]{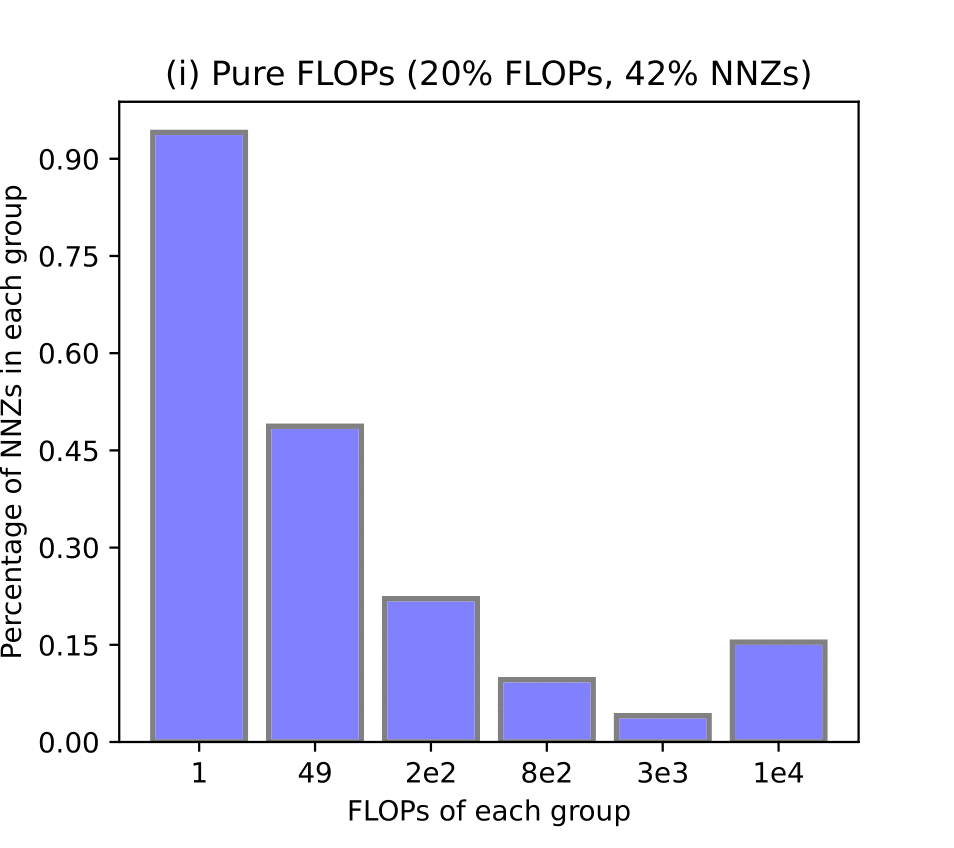}
    \end{minipage}%
    \begin{minipage}{0.48\linewidth}
        \includegraphics[width=0.95\columnwidth,trim=0.3cm 0.5cm 0.8cm 0.5cm]{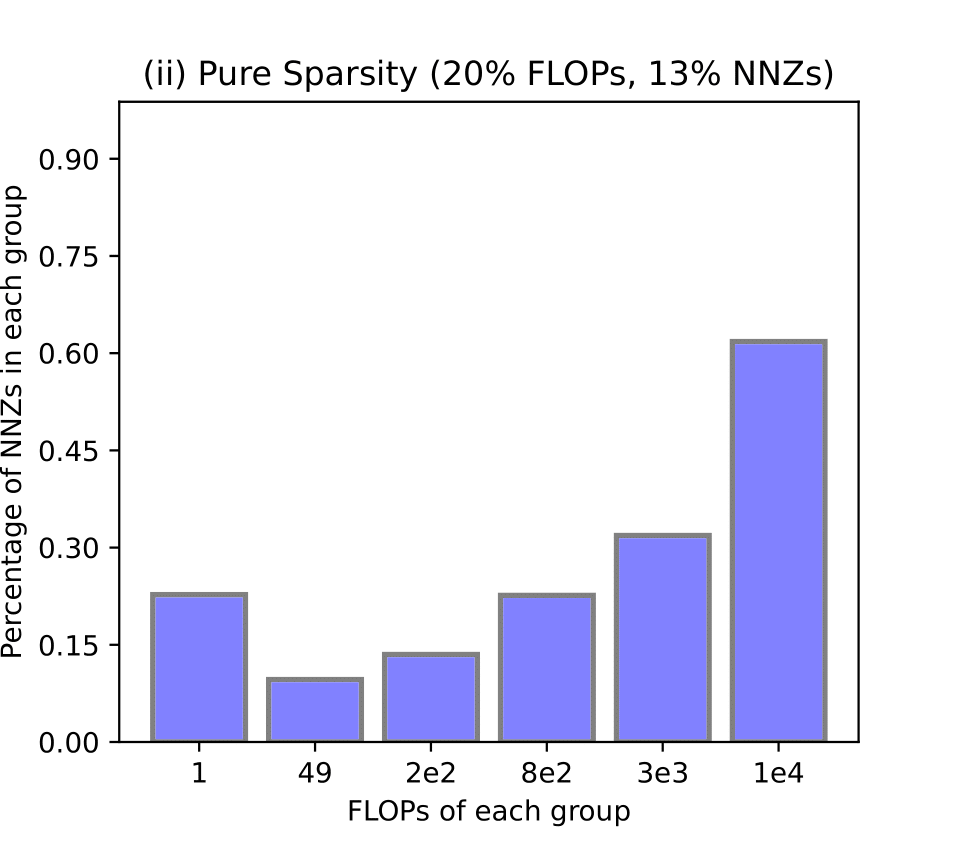}
    \end{minipage}%
    \hfill
    \begin{minipage}{0.48\linewidth}
        \includegraphics[width=0.95\columnwidth,trim=0.3cm 0.5cm 0.8cm 0.5cm]{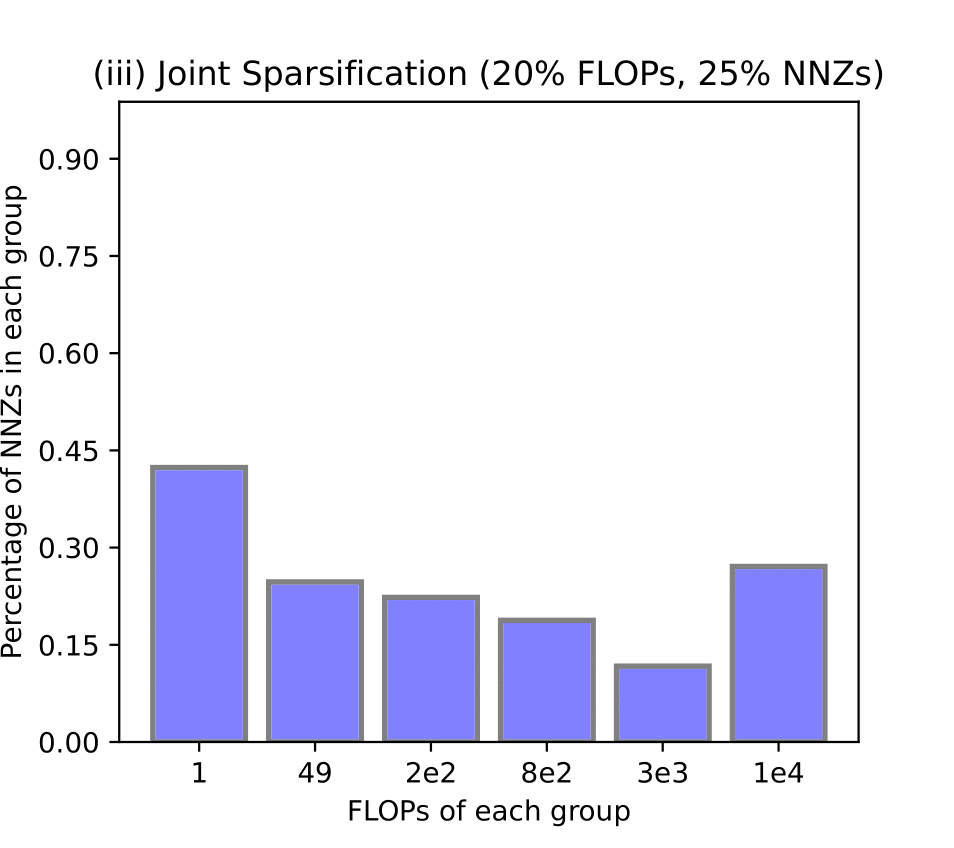}
    \end{minipage}%
    \begin{minipage}[c]{0.48\linewidth}
    \vspace{5mm}
       \caption{Sparsity of each group of Resnet50 pruned under a fixed FLOP budget (20\% of total FLOPs) by \modelname~across three scenarios: pure FLOP constraint, pure sparsity constraint, and joint sparsification. }
       \label{fig:bar}
    \end{minipage}%
    \vspace{0mm}
\end{figure}

It appears to us that the effectiveness of our approach arises from its ability to attain relatively uniform sparsity across the layers of the pruned network. In 
Figure~\ref{fig:bar}, we illustrate the sparsity of each group\footnote{As per Fact \ref{ass:layer}, parameters within the same layer have the same FLOPs cost and are categorized within the same group. Hence, we use group sparsity for a clearer comparison.} in the Resnet50 model, pruned under a fixed FLOP budget (20\% of total FLOPs) using (i) pure FLOP constraint, (ii) pure sparsity constraint, and (iii) joint sparsification. The joint sparsification approach strikes a nice balance between 
FLOP and NNZ constraints, leading to more even sparsity distribution across groups. This balanced sparsity distribution enhances accuracy, as demonstrated by \citet{Kusupati2020STR, singh2020woodfisher}.


\subsection{Gradual Pruning}\label{subsec:gradual_pruning}
In addition to evaluating our algorithms within the context of one-shot pruning methods, we have also integrated them into the gradual pruning framework~\citep{Gale-state-sparsity}. This framework employs a gradual, multi-step pruning process interspersed with Stochastic Gradient Descent (SGD) training epochs.  We compare our approach against  WF (with FLOP-aware pruning statistic)~\citep[Appendix S6]{singh2020woodfisher}, GMP~\citep{Gale-state-sparsity}, STR~\citep{Kusupati2020STR}, RIGL~\citep{evci2020rigging}, DNW~\citep{DNW2019}, and SNFS~\citep{dettmers2020sparse}. 
To match the FLOPs and NNZ of other baselines, we run \modelname~with various NNZ and FLOP budgets, yielding different accuracies for each budget.

\begin{table}[htbp]
\vspace{0mm}
\renewcommand{\arraystretch}{1}
\centering
\caption{\footnotesize Results of gradually pruning MobilenetV1 (top) and ResNet50 (bottom), comparing \modelname\ to other baselines.  \modelname~numbers are averaged over two runs. WF numbers are in the appendix of \citet{singh2020woodfisher}; numbers for other baselines are taken from \citet{Kusupati2020STR}.  }
\label{tab:gradual}
\resizebox{0.99\columnwidth}{!}
{\begin{tabular}{lcccc}
\toprule
                    & Sparsity &   FLOPs    & Top-1 Acc & NNZ \\
Method              & (\%) &                & (\%) &   \\ 
\midrule
MobilenetV1 & 0 (dense) & 569M & 71.95 & 4.21M \\
\midrule
GMP & 74.11 & 163M & 67.70 & 1.09M\\
STR & 75.28 & 101M & 68.35 & 1.04M\\
WF & 75.28 & 101M & 69.26 & 1.04M\\
\modelname & \textbf{75.28} & \textbf{101M} & \textbf{69.50} & \textbf{1.04M}\\
\hdashline
\noalign{\vskip 0.4ex}
WF & 75.28 & 92M & 68.69 & 1.04M\\
\modelname & \textbf{75.28} & \textbf{92M} & \textbf{69.22} & \textbf{1.04M}\\
\midrule
GMP & 89.03 & 82M & 61.80 & 0.46M\\
STR & 85.80 & 55M & 64.83 & 0.60M\\
\textbf{\modelname} & \textbf{90.00} & \textbf{55M} & \textbf{65.86} & \textbf{0.44M}\\
\hdashline
\noalign{\vskip 0.4ex}
STR & 90.00 & 40M & 61.51 & 0.44M\\
\textbf{\modelname} & \textbf{92.97} & \textbf{40M} & \textbf{61.75} & \textbf{0.30M}\\
\end{tabular}}
\resizebox{0.95\columnwidth}{!}{\begin{tabular}{lcccc}
\toprule
                    & Sparsity &   FLOPs    & Top-1 Acc & NNZ \\
Method              & (\%) &                & (\%) &   \\ 
\midrule
ResNet50 & 0 (dense) & 4.09G & 77.01 & 25.6M \\
\midrule
GMP & 90.00 & 409M & 73.91 & 2.56M \\
DNW & 90.00 & 409M & 74.00 & 2.56M \\
SNFS & 90.00 & 1.63G & 72.90 & 2.56M \\
SNFS + ERK & 90.00 & 960M & 72.90 & 2.56M \\
RigL & 90.00 & 515M & 72.00 & 2.56M \\
RigL + ERK & 90.00 & 960M & 73.00 & 2.56M\\
STR & 90.55 & 341M & 74.01 & 2.41M\\
WF & 90.23 & 335M & 74.34 & 2.49M\\
\modelname & \textbf{90.55} & \textbf{335M} & \textbf{74.72} & \textbf{2.41M}\\
\midrule
GMP & 95.00 & 204M & 70.59 & 1.28M \\
DNW & 95.00 & 204M & 68.30 & 1.28M \\
RigL* & 95.00 & 317M & 67.50 & 1.28M \\
RigL + ERK & 95.00 & $\sim$600M & 70.00 & 1.28M\\
STR & 95.03 & 159M & 70.40 & 1.27M \\
\modelname & \textbf{95.03} & \textbf{159M} & \textbf{71.81} & \textbf{1.27M} \\
\midrule
GMP & 98.00 & 82M & 57.90 & 0.51M \\
DNW & 98.00 & 82M & 58.20 & 0.51M \\
STR & 98.22 & 68M & 59.76 & 0.45M \\
\modelname & \textbf{98.22} & \textbf{68M} & \textbf{63.78} & \textbf{0.45M} \\
\bottomrule
\end{tabular}}
\vspace{0mm}
\end{table}


We carry out the gradual pruning evaluation on MobileNetV1 (4.2M parameters) and ResNet50 (25.6M parameters), and present the results in  Table \ref{tab:gradual}. We report the accuracy numbers of the competing methods from \citet{singh2020woodfisher} and \citet{Kusupati2020STR}---they report numbers with varying FLOP and sparsity configurations. To facilitate a fair comparison, we tune \modelname~so that it has the \textit{lowest} FLOPs and NNZs compared to the other methods --- we then observe that \modelname~still achieves superior accuracy compared to the others.  

We note that in gradual pruning, due to retraining, the fine-tuned model's accuracy closely mirrors that of the dense model. Hence, we do not expect a large absolute change in accuracy---this phenomenon is also observed in recent studies \citep{benbaki2023fast,singh2020woodfisher}. However, in terms of relative accuracy drop, \modelname~shows a significant improvement. Specifically, on a relative scale, the drop in accuracy of \modelname~(fine-tuned model versus the dense model) is $14\%\sim26\%$ smaller than that of WF and STR (our closest competitors) across different settings for Resnet50.

\section{Conclusions and Discussion}
In this work, we present an efficient optimization framework \modelname~for network pruning that considers both FLOP and sparsity constraints. Our algorithm builds on our custom efficient solver for approximately solving a structured integer linear program (ILP). By leveraging the low-rank structure of the optimization problem and employing advanced combinatorial optimization strategies, we significantly improve our algorithm in terms of both runtime and memory. Our experiments confirm that \modelname~outperforms existing pruning methods in performance under the same FLOP budgets, and its integration into gradual pruning frameworks leads to highly accurate networks with reduced FLOPs and NNZs.

\modelname~is capable of pruning networks to specified FLOP and NNZ budgets while maintaining accuracy (as much as possible), achieving a good balance between inference time and memory usage. This offers the practitioner to set precise compression targets to maximize resource usage for high-performing models. In the future, we seek to expand the FALCON framework to incorporate more flexible resource constraints, potentially enhancing its flexibility and effectiveness across various computing scenarios.


\section*{Acknowledgments}
This research is supported in part by grants from the Office of Naval Research (N000142112841 and N000142212665). We acknowledge the MIT SuperCloud and Lincoln Laboratory Supercomputing Center for providing HPC resources that have contributed to the research results reported within this paper. 
We thank Shibal Ibrahim for his helpful discussions.
\newpage
\bibliographystyle{plainnat}
\bibliography{references}
\newpage
\onecolumn
\appendix

\section{Proofs of Main Theorems in  Section~\ref{sect:impact} and Auxiliary Results}\label{app:proofs}
\subsection{Proof of Theorem \ref{thm:complexity}}\label{subsec:prooftime}

\begin{proof}

Our proof centers around the time complexity for each iteration of Algorithm \ref{alg:proj}. In each iteration, the algorithm computes the value of

\begin{equation}\label{eq:dualcald}
g(\lambda_2)=\min_{\lambda_1\ge 0} D(\lambda_1,\lambda_2) = D\big( \max\{(I-\lambda_2f)_{(S)},0\},\lambda_2\big)
\end{equation}
for a given $\lambda_2\ge 0$. It is clear that the primary computational expense lies in determining $(I-\lambda_2f)_{(S)}$, which represents the $S$-th largest element of $I-\lambda_2f$. Utilizing conventional algorithms (e.g., quicksort) for calculating $(I-\lambda_2f)_{(S)}$ requires $O(p)$ time. However, we will demonstrate that the time complexity can be reduced by leveraging the structure of vector $f$.

Notably, Fact \ref{ass:layer} states that parameters within the same group have identical FLOP costs. This implies that the size relationship between $(I-\lambda_2 f)_i$ and $(I-\lambda_2 f)_{i'}$ remains constant irrespective of the value of $\lambda_2$, provided $i$ and $i'$ belong to the same group $C_j$ (for some $j\in[L]$). Hence, the values in each group can be pre-sorted such that for any value of $\lambda_2$, the elements in $(I-\lambda_2f)$ can be divided into $L$ sorted arrays without additional computational cost. 

We propose Algorithm \ref{alg:sort2}, which capitalizes on this structure to find the $S$-th largest element of $(I-\lambda_2f)$ efficiently. The core idea of Algorithm \ref{alg:sort2} is to utilize the structure of sorted arrays to partition $I-\lambda_2f$ based on a pivot element efficiently. This allows for recursively minimizing the search space to the desired $S$-th largest element by removing the unnecessary portions of the array. In this algorithm, the pivot element $a_{\sigma_t}$ is wisely chosen to ensure that the number of elements in the search space is reduced by a constant factor at each recursive step.

It is straightforward to verify that Algorithm \ref{alg:sort2} consistently maintains the desired $S$-th largest element in the search space, thereby assuring its correctness. Now, we  provide a formal estimation of the time complexity of Algorithm \ref{alg:sort2}. Note that sorting a vector of length $p$ requires $O(p\log p)$ time, and partitioning a sorted array of length $p$ based on a value takes $O(\log p)$ time. The time complexity for line 1-7 in Algorithm \ref{alg:sort2} is thus
\begin{equation}
O(L\log L) + O\left(\sum_{j=1}^L \log(|v_j|)\right) \le O(L\log L)+O(L\log p) = O(L\log p),
\end{equation}
 Next, we present the following lemma:
\begin{lemma}\label{lem:length}
The arrays $\{u_j\}_{j=1}^L$ and $\{s_j\}_{j=1}^L$ generated in Algorithm \ref{alg:sort2} satisfies 
\begin{equation}
\sum_{j=1}^L|u_j| \ge p/4,\,\, \sum_{j=1}^L|s_j| \ge p/4.
\end{equation}
\end{lemma}
\begin{proof}
From the fact that $a_{\sigma_1} \ge a_{\sigma_2}\ge\cdots\ge a_{\sigma_L}$ and the definition of $t$ and $\{u_j\}_{j=1}^L$, we derive
\begin{equation}
\begin{aligned}
\sum_{j=1}^L|u_j| &= \left| \left\{ a\in \cup_{j=1}^L \{v_j\}\big| a\ge a_{\sigma_t}  \right\} \right| \\
&\ge \sum_{j=1}^t \left| \left\{ a\in \ v_{\sigma_j}\big| a\ge a_{\sigma_j}  \right\} \right|\\
& \ge \frac{1}{2}\sum_{j=1}^t|v_{\sigma_j}|\ge p/4.
\end{aligned}
\end{equation}
Likewise, given $a_{\sigma_1} \ge a_{\sigma_2}\ge\cdots\ge a_{\sigma_L}$ and the definition of $t$ and $\{s_j\}_{j=1}^L$, we can deduce that
\begin{equation}
\begin{aligned}
\sum_{j=1}^L|s_j| &= \left| \left\{ a\in \cup_{j=1}^L \{v_j\}\big| a\le a_{\sigma_t}  \right\} \right| \\ 
& \ge \sum_{j=t}^L \left| \left\{ a\in \ v_{\sigma_j}\big| a\le a_{\sigma_j}  \right\} \right| \\
& \ge \frac{1}{2}\sum_{j=t}^L|v_{\sigma_j}|= \frac{p}{2}-\frac{1}{2}\sum_{j=1}^{t-1}|v_{\sigma_j}|\ge p/4.
\end{aligned}
\end{equation} 
This concludes the proof.
\end{proof}
From Lemma \ref{lem:length}, it follows that the number of elements remaining in sorted arrays decreases by at least $1/4$ after each recursion. Therefore, Algorithm \ref{alg:sort2} is able to find the $S$-th largest element in at most $O(\log p)$ recursive steps. Hence, the overall time complexity for Algorithm \ref{alg:sort2} is $O\big(L(\log p)^2\big)$.

Once we have computed the $(I-\lambda_2 f)_{(S)}$, $D\big( \max\{(I-\lambda_2f)_{(S)},0\},\lambda_2\big)$ in \eqref{eq:dualcald} can be calculated in $O(L\log p)$ time if we pre-compute and store the cumulative sum of $L$ sorted arrays. Therefore, $g(\lambda_2)$ can be evaluated in $O\big(L(\log p)^2\big)$ time for any given $\lambda_2$.

Finally, we analyze the convergence of Algorithm \ref{alg:proj}. The algorithm employs the Golden-section search method, which, as described in \cite[Chapter 10]{yao2007early}, guarantees that $\lambda_2^{max}$ and $\lambda_2^{min}$ consistently provide a valid upper and lower bound for the optimal value $\lambda_2^* = \argmin_{\lambda_2\ge 0} g(\lambda_2)$. At each iteration, the difference between the bounds, $\lambda_2^{max} - \lambda_2^{min}$, is multiplied by a factor of $(\sqrt{5} - 1) / 2$. Consequently, it takes $O(\log (1/\varepsilon))$ iterations to obtain an $\varepsilon$-accurate solution. To conclude, the overall time complexity for Algorithm \ref{alg:proj} to get a $\varepsilon$-accurate solution is $O\big(p\log p +L (\log p)^2 \log (1/\varepsilon)\big)$, where the term $O(p\log p)$ comes from pre-processing.

\begin{algorithm}[h]
\begin{algorithmic}[1]
\REQUIRE $L$ sorted arrays $v_1,v_2,\dots,v_L$ with length denoted as $|v_1|,|v_2|,\dots,|v_L|$, and an integer $S \in \left[1,p:=\sum_{j=1}^L |v_i|\right]$
\STATE Let $a_j$ be the median of array $v_j$,\,$\forall j\in [L]$.
\STATE Sort $\{a_{j}\}_{j=1}^L$ to obtain $a_{\sigma_1} \ge a_{\sigma_2}\ge\cdots\ge a_{\sigma_L}$. 
\STATE Let $t=\argmin\left\{t\in [L]\big| \sum_{j=1}^t|v_{\sigma_j}|\ge p/2\right\}$.
\FOR{$j=1,2,\dots,L$}
\STATE Split $v_j$ into two sort arrays $u_j,\,u_j'$ such that $a\in v_j$ belongs to $u_j$ if and only if $a\ge a_{\sigma_t}$.
\STATE Split $v_j$ into two sort arrays $s_j,\,s_j'$ such that $a\in v_j$ belongs to $s_j$ if and only if $a\le a_{\sigma_t}$.
\ENDFOR
\IF{$\sum_{j=1}^L|u_j| \le S-1$}
\STATE \textbf{Return} \textsf{Largesort}$(\{u_j'\}_{j=1}^L,S-\sum_{j=1}^L|u_j|)$
\ELSIF{$\sum_{j=1}^L|s_j| \ge p-S$}
\STATE \textbf{Return} \textsf{Largesort}$(\{s_j'\}_{j=1}^L,S)$
\ELSE
\STATE \textbf{Return} $a_{\sigma_t}$
\ENDIF
\end{algorithmic}
\caption{\,\,\textsf{Largesort}$(\{v_j\}_{j=1}^L,S)$: Calculate the $S$-th largest element in $L$ sorted arrays}
\label{alg:sort2}
\end{algorithm}

\end{proof}

\subsection{Proof of Theorem \ref{thm:optgap}}
\begin{proof}
Given the optimal dual solution, $(\lambda_1^*, \lambda_2^*)$, we first discuss how to obtain an optimal primal solution for the relaxed linear programming problem expressed in \eqref{eq:lp-impact}.

Note that $\lambda_1$ and $\lambda_2$ are Lagrangian multipliers associated with the constraints $\sum_{i=1}^p z_i\le S$ and $\sum_{i=1}^p f_iz_i\le F$, respectively. If $\lambda_2^* = 0$, then the constraint $\sum_{i=1}^p f_iz_i \le F$ is inactive, allowing us to disregard this constraint in the primal problem and solve it by sorting $I_i$'s to obtain the primal optimal solution $z^*$. If $\lambda_1^* = 0$,  then the constraint $\sum_{i=1}^p z_i \le S$ is inactive. In this case, we may sort \( I_i/f_i \) to get the optimal primal solution $z^*$. 

We now consider the case where both $\lambda_1^*$ and $\lambda_2^*$ are greater than $0$. The KKT condition implies that the primal optimal solution $z^*$ satisfies:

\begin{equation}\label{eq:lp-kkt}
    \begin{aligned}
        z^*_i &\in \left\{
        \begin{array}{ll}
           \{1\},  &  \text{if  } i\in \mathcal{Z}_1:=\left\{i\in [p]\big| I_i-\lambda_1^*-f_i\lambda_2^* > 0\right\}, \\
            \lbrack0,1 \rbrack, &\text{if  } i\in \mathcal{Z}_2:=\left\{i\in [p]\big| I_i-\lambda_1^*-f_i\lambda_2^* = 0\right\}, \\
            \{0\}, &  \text{if  } i\in \mathcal{Z}_3:=\left\{i\in [p]\big| I_i-\lambda_1^*-f_i\lambda_2^* < 0\right\}, \\
        \end{array}
        \right. \\
        0&= \lambda^*_1(S - \sum_{i=1}^p z_i^*), \\
        0&=\lambda^*_2(F - \sum_{i=1}^p f_iz_i^*). \\
    \end{aligned}
\end{equation}

From this, we can conclude that $ \sum_{i=1}^p z_i^*=S$ and $\sum_{i=1}^p f_iz_i^*=F$ hold. It remains to determine the values of $z^*_i$ for $i \in \mathcal{Z}_2$. We fix $z_i^*=1$ for $i\in \mathcal{Z}_1$ and $z_i^*=0$ for $ \mathcal{Z}_3$ in \eqref{eq:lp-impact}, and solve the resulting problem

\begin{equation}
\begin{aligned}
\max_{z}\,\,\,\, \,\,& \sum_{i\in \mathcal{Z}_2} I_i z_i, \\
\text{s.t.}\,\,\,\,\,\,&   \sum_{i\in \mathcal{Z}_2} f_iz_i = F-\sum_{i\in \mathcal{Z}_1} f_i,\,\,\, \sum_{i\in \mathcal{Z}_2} z_i = S-|\mathcal{Z}_1|,\\ 
&z_i\in [0,1]\,\,\,\forall i\in [p].
\end{aligned}    
\end{equation}

Since $I_i-\lambda_1^*-f_i\lambda_2^* = 0$ for any $i \in \mathcal{Z}_2$, the above problem is equivalent to the following feasibility problem and can be readily solved.

\begin{equation}
\begin{aligned}
\max_{z}\,\,\,\, \,\,&  \sum_{i\in \mathcal{Z}_2} I_i z_i=\lambda_2^* (F-\sum_{i\in \mathcal{Z}_1} f_i)+ \lambda_1^*( S-|\mathcal{Z}_1|), \\
\text{s.t.}\,\,\,\,\,\,&   \sum_{i\in \mathcal{Z}_2} f_iz_i = F-\sum_{i\in \mathcal{Z}_1} f_i,\,\,\, \sum_{i\in \mathcal{Z}_2} z_i = S-|\mathcal{Z}_1|,\\ 
&z_i\in [0,1]\,\,\,\forall i\in [p].
\end{aligned}    
\end{equation}

Given an optimal primal solution $z^*$ of the relaxed problem \eqref{eq:lp-impact}, we now focus on retrieving a feasible solution $\hat z$ for the MIP problem \eqref{eq:main-impact} and analyzing its quality. Denote $\mathcal{Z}_1^*:=\{i| z_i^*=1\}$, $\mathcal{Z}_2^*:=\{i| z_i^*\in (0,1)\}$, and $\mathcal{Z}_3^*:=\{i| z_i^*=0\}$.The KKT condition \eqref{eq:lp-kkt} implies that $I_i=\lambda_1^*+f_i\lambda_2^*$ for $i\in \mathcal{Z}_2^*$.

Recall that in Assumption \ref{ass:layer}, all parameters are partitioned into $L$ groups $C_1,\dots,C_L$. Suppose for some $j\in [L]$ there exists $i, i' \in C_j\cap \mathcal{Z}^*_2 $ with $i\neq i'$. From the KKT condition and $f_i=f_{i'}$, we have

\begin{equation}
    I_i= \lambda_1^* + f_i\lambda_2^* = \lambda_1^* + f_{i'}\lambda_2^* = I_{i'}.
\end{equation}

Therefore, we can replace the value of $z^*_i$ and $z^*_{i'}$ with $\min\{1, z^*_i + z^*_{i'}  \}$ and $\max\{0, z^*_i + z^*_{i'} - 1  \}$, without altering the objective or violating the constraints. This implies that after adjustment, $z^*$ is still an optimal primal solution. By adjusting $z^*$, we can reduce the size of $|C_j\cap \mathcal{Z}^*_2|$ as long as it is no less than $2$. Therefore, without loss of generality, we assume that $|C_j\cap \mathcal{Z}^*_2|\le 1$ for any $j\in[L]$. Now we set 

\begin{equation}
    \hat z_i = \left\{
        \begin{array}{ll}
           1 & \text{if  } i\in \mathcal{Z}^*_1,\\
           0 & \text{if  } i\in \mathcal{Z}^*_2 \cup \mathcal{Z}^*_3.\\
        \end{array}\right.
\end{equation}

Since $\sum_{i=1}^p \hat z_i\le \sum_{i=1}^p z_i^*\le S$ and $\sum_{i=1}^p f_i\hat z_i\le \sum_{i=1}^p f_iz_i^*\le F$, $\hat z$ is a feasible solution to the MIP problem \eqref{eq:main-impact}. On the other hand, since $z^*$ is the optimal solution of the relaxed problem of \eqref{eq:main-impact}, $Q_I(z^*)=\sum_{i=1}^p I_iz_i$ provides an upper bound of $Q^*_I$, the optimal objective of the MIP problem. Together with $D(\lambda_1^*,\lambda_2^*) = \sum_{i=1}^pI_i z_i^*\ge S\lambda_1^*+F\lambda_2^*$, we conclude that

\begin{equation}
    \begin{aligned}
        \frac{Q_I^*-Q_I(\hat z)}{Q_I^*} & \le \frac{\sum_{i=1}^p I_i (z_i^*-\hat z_i)}{\sum_{i=1}^p I_i z_i^*}, \\        
        & \le \frac{\sum_{i\in \mathcal{Z}^*_2} I_i}{D(\lambda_1^*,\lambda_2^*)},\\
        & =  \frac{\sum_{i\in \mathcal{Z}^*_2} \lambda_1^* + f_{i}\lambda_2^* }{D(\lambda_1^*,\lambda_2^*)},\\
        & \le  \frac{\sum_{j=1}^L \lambda_1 + \sum_{j=1}^L f^j \lambda_2^* }{D(\lambda_1^*,\lambda_2^*)},\\
        &\le  \frac{L\lambda_1^* + L_f\lambda_2^* }{S\lambda_1^*+F\lambda_2^*},\\
        &\le \max\left\{\frac{L}{S},\frac{L_f}{F}\right\},
    \end{aligned}
\end{equation}
which completes the proof.
\end{proof}

\subsection{Auxiliary results} \label{sect:auxthm}

In this subsection, we present Proposition \ref{prop:mpeq}, which establishes the equivalence between magnitude pruning and solving Problem \eqref{eq:MP}, along with the proof of Lemma \ref{lemma:proj}.

\subsubsection{Equivalence between magnitude pruning and Problem \eqref{eq:MP}}
\begin{proposition}\label{prop:mpeq}
Using magnitude pruning to prune the weight vector $\bar w$ with NNZ budget $S$ is equivalent to solving the following cardinality-constrained problem
\begin{equation}\label{eq:MP2}
\max_{z\in\{0,1\}^p}~~ Q_I(z):=\sum_{i=1}^p I_i z_i, ~~
\text{s.t.}~~\sum_{i=1}^p z_i \le S.
\end{equation}
with $I_i=(\bar w_i)^2$ and setting the pruned weights as $w_i = \bar w_iz_i$ after Problem~\eqref{eq:MP} is solved for $z_i$'s.
\end{proposition}
\begin{proof}
    MP selects a set of parameters with the least absolute values to remove while adhering to NNZ budget. Let $\mathcal{I}$ denote the set of indices of the $S$ largest elements in $(\bar w)^2$. For simplicity, we assume that such a set is unique; other cases can be handled similarly. Applying MP to prune the weight vector $\bar w$ with NNZ budget $S$ results in a vector $w$ expressed as
    \begin{equation}
        w_i = \left\{\begin{array}{ll}
            \bar w_i & \text{ if }i\in \mathcal{I}, \\
            0  & \text{ otherwise}.
        \end{array}\right.
    \end{equation}
On the other hand, $\{z_i\}_{i=1}^p$ are binary variables and $I_i\ge 0$ in Problem \eqref{eq:MP2}. Given the NNZ budget $\sum_{i=1}^p z_i \le S$, the optimal solution of  Problem \eqref{eq:MP2} is 
\begin{equation}
    z_i = \left\{\begin{array}{ll}
        1 & \text{ if } i\in \mathcal{I}, \\
        0  & \text{ otherwise}.
    \end{array}\right.
\end{equation}
    Therefore, setting $w_i = \bar w_iz_i$ results in a pruned weight vector identical to that produced by MP.
\end{proof}

\subsubsection{Proof of Lemma \ref{lemma:proj}}
\begin{proof}
By setting $z_i=\mathbf{1}_{x_i\neq 0}$, the projection problem \eqref{eq:dfoproj} can be reformulated as
\begin{equation}\label{eq:lem411}
    \begin{aligned}
        \argmin_{x\in\mathbb{R}^p,z\in\{0,1\}^p}~~~&  \sum_{i=1}^p (x_i-\bar x_i)^2, \\
        \text{s.t.}~~~& \sum_{i=1}^p z_i\le S,~~ \sum_{i=1}^p f_iz_i\le F.
    \end{aligned}
\end{equation}
Note that once $\{z_i\}_{i=1}^p$ are fixed, the optimal choices of $\{x_i\}_{i=1}^p$ that minimize the objective are given by $x_i=\bar x_i z_i,~\forall i\in[p]$. Therefore, we can replace $x_i$ with $\bar x_i z_i$ in \eqref{eq:lem411}, which leads to the following equivalent problem:
\begin{equation}\label{eq:lem412}
    \begin{aligned}
        \argmin_{z\in\{0,1\}^p}~~~&  \sum_{i=1}^p (\bar x_i z_i-\bar x_i)^2 = \sum_{i=1}^p (\bar x_i)^2(2- 2z_i), \\
        \text{s.t.}~~~& \sum_{i=1}^p z_i\le S,~~ \sum_{i=1}^p f_iz_i\le F.
    \end{aligned}
\end{equation}
This problem is further equivalent to \eqref{eq:main-impact} with $I_i=(\bar x_i)^2~\forall i\in[p]$. Hence, solving \eqref{eq:main-impact} with $I_i=(\bar x_i)^2~\forall i\in[p]$ and setting $x_i=\bar x_i z_i,~\forall i\in[p]$ provides a solution to the projection problem \eqref{eq:dfoproj}. 
\end{proof}

\section{Details on Optimization Formulation and Algorithm}
\subsection{Optimization formulation details}\label{subapp:formulation-details}
Our optimization-based framework use a local model $\mathcal{L}$ around the pre-trained weights $\bar w$:
\begin{equation}
\mathcal{L}(w)=\mathcal{L}(\bar w)+\nabla \mathcal{L}(\bar w)^\top (w-\bar w)+ \ \frac12(w-\bar w)^\top \nabla^2\mathcal{L}(\bar w)(w-\bar w)+O(\|w-\bar w\|^3).
\end{equation}
By selecting appropriate gradient and Hessian approximations $g\approx\nabla\mathcal{L}(\bar w), H\approx\nabla^2\mathcal{L}(\bar w)$ and disregarding higher-order terms, we derive $Q_{L_0}(w)$ as a local approximation of the loss $\mathcal{L}$:
\begin{equation}
Q_{L_0}(w) :=\mathcal{L}(\bar w)+ g^\top(w-\bar w)+\frac12(w-\bar w)^\top H(w-\bar w).
\end{equation}

\paragraph{Choices of Hessian approximation $H$.}
Previous works~\citep{hassibi1992second,singh2020woodfisher,benbaki2023fast} have approximated the Hessian matrix using the empirical Fisher information matrix derived from $n$ samples. It has been found that using a few hundred to one thousand samples is sufficient for Hessian estimation, resulting in a low-rank representation:
\begin{equation}\label{eq:fisherhess}
\widehat H=\frac1n\sum_{i=1}^n\nabla \ell_i(\bar w)\nabla \ell_i(\bar w)^\top=\frac1n X^\top X\in\R^{p\times p},
\end{equation}
where $X=[\nabla \ell_1(\bar w),\ldots, \nabla \ell_n(\bar w)]^\top\in\R^{n\times p}$. This low-rank structure is leveraged in \citep{benbaki2023fast} to circumvent the need for storing the full dense Hessian and to accelerate convergence.

\paragraph{Choices of gradient approximation $g$.}
Traditional pruning methods for neural networks typically assume that the pre-trained weights $\bar w$ represent a local optimum of the loss function $\mathcal{L}$ and thus set the gradient $g$ to zero. However, in practice, the gradient of a pre-trained neural network's loss function may not be zero due to early stopping or approximate optimization~\citep{yao2007early}. Additionally, pruning the local quadratic model at a general reference point, where the gradient is not zero, may yield more desirable results. Consequently, we propose approximating the gradient using the stochastic gradient, which leverages the same samples as those used for estimating the Hessian:
\begin{equation}\label{eq:grad-approx}
g=\frac1n\sum_{i=1}^n\nabla \ell_i(\bar w)=\frac1n X^\top e\in\R^p,
\end{equation}
where $e\in\R^n$ is a vector of all ones.

\paragraph{Objective construction}

Utilizing the Hessian approximation \eqref{eq:Hessian} and gradient approximation \eqref{eq:grad-approx}, we formulate the network pruning as an optimization problem that minimizes $Q_{L_0}(w)$ while adhering to NNZ and FLOPs constraints:
\begin{equation}\label{eq:miqp}
\min_w ~~ Q_{L_0}(w)
\qquad \text{s.t. }~~~\|w\|_0\le S,\,\,\,\,\|w\|_{0,f}\le F.
\end{equation}

Based on our empirical observations, the performance of the pruned model is heavily dependent on the accuracy of the quadratic approximation $Q_{L_0}(w)$ for the loss function. As this approximation is local, it is imperative to ensure that the weights $w$ during pruning remain in close proximity to the initial weights $\bar{w}$.  Thus, we incorporate a ridge-like regularizer of the form $\|w-\bar w\|^2$ into the objective in~\eqref{eq:miqp}, resulting in the following problem:
\begin{equation}\label{eq:miqp2}
\begin{aligned}
\min_w ~~~ &Q_{L}(w):= g^\top(w-\bar w)+\frac12(w-\bar w)^\top H(w-\bar w) + \frac{n\lambda}{2}\|w-\bar w\|^2,\\
\text{s.t.}~~~&\|w\|_0\le S\,\,\,\,\|w\|_{0,f}\le F,
\end{aligned}
\end{equation}
where $\lambda\geq 0$ is a parameter governing the strength of the regularization.

\paragraph{Special consideration for block approximation}


We employ a block-wise approximation $\widehat H_B$ of $\widehat H$, considering only limited-size blocks on the diagonal of $\widehat H$ and disregarding off-diagonal elements.We treat the set of variables corresponding to a single layer in the network as a block and uniformly subdivide these blocks such that the size of each block does not exceed a given parameter $B_{size}$.  Based on our empirical observations, this block-wise approximation $\widehat H_B$  delivers a more accurate approximation of the Hessian matrix when compared to the original $\widehat H$. 

Furthermore, we multiply the block-wise approximation $\widehat H_B$ by a constant factor $\rho$, ensuring that $\widehat H \preceq H:= \rho\widehat H_B$. This guarantees that $Q_{L_0}(w)$ serves as an (approximate) upper bound of $L(w)$:
\begin{equation}\label{eq:defrho}
\begin{aligned}
L(w) &\approx \mathcal{L}(\bar w)+ g^\top(w-\bar w)+\frac12(w-\bar w)^\top\widehat H(w-\bar w) \\
&\le \mathcal{L}(\bar w)+ g^\top(w-\bar w)+\frac\rho 2(w-\bar w)^\top\widehat H_B(w-\bar w) = Q_{L_0}(w).
\end{aligned}
\end{equation}
This procedure is inspired by the well-known majorization-minimization approach~\citep{lange2016mm}, where one minimizes an upper bound on the objective at each iteration. In our experiments, we observe that this approach can help mitigate the effects of imprecise Hessian and gradient approximations during the optimization procedure.

Given a disjoint partition $\{\mathcal{B}_j\}_{j=1}^K$ of $\{1,2,\dots,p\}$ and assuming blocks of size $|\mathcal{B}_1|\times |\mathcal{B}_1|,\dots,|\mathcal{B}_K|\times |\mathcal{B}_K|$ along the diagonal of $\widehat H_B$, we can represent $H$ as:
\begin{equation}\label{eq:hessfree}
H=\rho \widehat H_B = \rho \diag(X_{\mathcal{B}_1}^\top X_{\mathcal{B}_1},\dots, X_{\mathcal{B}_K}^\top X_{\mathcal{B}_K}),
\end{equation}
where $X_{\mathcal{B}_j}$ is a submatrix of $X=[\nabla \ell_1(\bar w),\ldots, \nabla \ell_n(\bar w)]^\top$ with columns in $\mathcal{B}_j$.
Consequently, our optimization formulation \eqref{eq:miqp2} is Hessian-free, necessitating only the storage of a matrix $X\in \mathbb{R}^{n\times p}$. Moreover, by leveraging the representation \eqref{eq:hessfree}, the matrix-vector multiplications involving $H$ in the DFO update
\begin{equation}
    \begin{aligned}
         w^{t+1} &= \textsf{DFO}(w^t,\tau^s) := P_{S,F}\left(w^t-\tau^s \nabla Q_L(w^t)\right)\\
         =&P_{S,F}\left(w^t-\tau^s \big(g+ H(w^t-\bar w)+n\lambda(w^t-\bar w)\big)\right), 
    \end{aligned}
\end{equation}
can be executed by operating on the matrix $X\in \mathbb{R}^{n\times p}$ with cost $O(np)$.

To conclude, our algorithm operates solely on the low-rank matrix $X$ and circumvents the need to directly store and solve problem \eqref{eq:miqp2}, which would be computationally demanding due to the considerable size of the $p \times p$ matrix $H$ for large networks. Our approach results in significant improvements in both memory usage and runtime.

\subsection{Algorithmic details}\label{subapp:algorithm-details}

\subsubsection{Active set updates}\label{subsec:active-set}

The active set strategy has demonstrated its effectiveness in various contexts by reducing complexity~\citep{nocedal1999numerical,hazimeh2020fast}. This section demonstrates how to apply an active set strategy to accelerate our DFO updates. We begin by projecting the pre-trained weight onto a set with NNZ budget $S' \ge S$ and FLOPs budget $F' \ge F$, obtaining $\hat w=P_{S',F'}(\bar w)$. We then define the initial active set as $\mathcal{Z}=\supp(\hat w):=\{i| \hat w_i\neq 0\}$.

During each iteration, we limit DFO updates to the current active set $\mathcal{Z}$. Upon convergence, we execute a single DFO update on the entire vector to identify an improved solution $w$ with $\supp(w) \not\subseteq \mathcal{Z}$. If no such $w$ exists, the algorithm terminates; otherwise, we update $\mathcal{Z}\gets \mathcal{Z}\cup \supp(w)$ and repeat the process. Algorithm \ref{alg:active} offers a detailed depiction of the active set method.

\begin{algorithm}[h]
\begin{algorithmic}[1]
\REQUIRE Initial solution $w^0$, stepsize $\tau^s$, the number of DFO iterations $T$, and an initial set $\mathcal{Z}^0$.
\FOR {$t=0,1,\ldots$}
\FOR {$t'=0,1,\ldots, T$}
\STATE Perform DFO update $w^{t}|_{\mathcal{Z}^t} = \textsf{DFO}(w^t|_{\mathcal{Z}^t},\tau^s)$ restricted on $\mathcal{Z}^t$ 
\ENDFOR
\STATE Find $\tau'$ via line search such that $w^{t+1}=\textsf{DFO}(w^{t},\tau')$ satisfies\\ \qquad(i)~$Q_L(w^{t+1})<Q_L(w^{t})$ ~(ii)~ $\supp(w^{t+1})\not\subseteq  \mathcal{Z}^t$ 
\IF {such $\tau'$ does not exist}
\STATE \textbf{break}
\ELSE 
\STATE $\mathcal{Z}^{t+1}\gets \mathcal{Z}^t\cup \supp(w^{t+1})$ 
\ENDIF
\ENDFOR
\end{algorithmic}
\caption{Active set with DFO: \texttt{ACT-DFO}$(w^0,\tau^s,T,\mathcal{Z}^0)$}
\label{alg:active}
\end{algorithm}

\subsubsection{Backsolve via Woodbury formula}\label{subsec:backsolve}
As the problem's dimensionality increases, DFO updates becomes increasingly computationally demanding, even when employing an active set strategy. To address this challenge, we propose a backsolve approach that reduces complexity while maintaining a slightly suboptimal solution. The backsolve approach computes the optimal solution precisely on a restricted set. We initially apply DFO updates several times to obtain a feasible solution $w$ and then restrict the problem to the set $\mathcal{Z}:= \text{supp}(w)$. Under this restriction, problem \eqref{eq:miqp2} simplifies to a quadratic problem without any constraint, and its minimizer is given by:
\begin{equation}\label{eq:bs}
    w^*_{\mathcal{Z}}= (n\lambda I+H_{\mathcal{Z}})^{-1}\big( (H_{\mathcal{Z}}+ n\lambda I)\bar w_{\mathcal{Z}} + g_{\mathcal{Z}}\big),
\end{equation}
where $H_\mathcal{Z}\in \mathbb{R}^{|\mathcal{Z}|\times |\mathcal{Z}|}$ denotes a submatrix of $H$ with rows and columns only in $\mathcal{Z}$.

Recall that $\{\mathcal{B}_j\}_{j=1}^K$  is a disjoint partition of $\{1,2,\dots,p\}$, and  $H=\rho \widehat H_B$ is a block diagonal matrix with block sizes $|\mathcal{B}_1|\times |\mathcal{B}_1|,\dots,|\mathcal{B}_K|\times |\mathcal{B}_K|$ along the diagonal, represented as:
\begin{equation}
H=\rho \widehat H_B = \rho \diag(X_{\mathcal{B}_1}^\top X_{\mathcal{B}_1},\dots, X_{\mathcal{B}_K}^\top X_{\mathcal{B}_K}),
\end{equation}
where $X_{\mathcal{B}_j}$ is a submatrix of $X=[\nabla \ell_1(\bar w),\ldots, \nabla \ell_n(\bar w)]^\top$ with columns in $\mathcal{B}_j$. Based on this representation, $H_\mathcal{Z}$ can be expressed as
\begin{equation}\label{eq:hzlow}
H_\mathcal{Z}= \rho \diag(X_{\mathcal{B}_1'}^\top X_{\mathcal{B}_1'},\dots, X_{\mathcal{B}_K'}^\top X_{\mathcal{B}_K'}),
\end{equation}
where $\mathcal{B}_j'=\mathcal{B}_j\cap \mathcal{Z},\,\forall j\in [K]$.

By leveraging the low-rank representation \eqref{eq:hzlow} and utilizing  Woodbury formula~\citep{max1950inverting}, one can compute each sub-vector $w_{\mathcal{B}_j'}^*$ of $w_\mathcal{Z}^*$ in ~\eqref{eq:bs} using matrix-vector multiplications involving only $X_{\mathcal{B}_j'}$ (or its transpose) and one matrix-matrix multiplication via
\begin{equation}\label{eqn:bs-detail}
\begin{aligned}
    w_{\mathcal{B}_j'}^*&= (n\lambda I+H_{\mathcal{B}_j'})^{-1}\big( (H_{\mathcal{B}_j'}+ n\lambda I)\bar w_{\mathcal{B}_j'} + g_{\mathcal{B}_j'}\big) \\
    &=(n\lambda)^{-1}[ I-\rho X_{\mathcal{B}_j'}^\top(n\lambda I+\rho X_{\mathcal{B}_j'}X_{\mathcal{B}_j'}^\top)^{-1}X_{\mathcal{B}_j'}]\cdot\big( n\lambda\bar w_{\mathcal{B}_j'} + \rho X_{\mathcal{B}_j'}^\top X_{\mathcal{B}_j'}\bar w_{\mathcal{B}_j'} + g_{\mathcal{B}_j'}\big).
\end{aligned}
\end{equation}
It takes $O(n^3+n^2|\mathcal{B}_j'|)$ operations to compute $w_{\mathcal{B}_j'}^*$, thus the overall complexity for backsolve is $\sum_{j=1}^K O(n^3+n^2|\mathcal{B}_j'|)=O(n^3K+n^2S)$. Our proposed backsolve procedure is detailed in Algorithm \ref{alg:bs}. 

\begin{algorithm}[h]
\begin{algorithmic}[1]
\REQUIRE Initial solution $w^0$.
\STATE Construct an initial active set $\mathcal{Z}^0$; determine stepsize $\tau^s$ and the number of DFO iterations $T$
\STATE Set $w=\texttt{ACT-DFO}(w^0,\tau^s, T, \mathcal{Z}^0)$
\STATE $\mathcal{Z}\gets \supp(w)$
\FOR{$j=1,\dots,K$}
\STATE Set $\mathcal{B}_j'=\mathcal{B}_j\cap \mathcal{Z}$ and compute $w_{\mathcal{B}_j'}^*$ according to \eqref{eqn:bs-detail}.
\ENDFOR
\STATE Concatenating sub-vectors $\{w_{\mathcal{B}_j'}^*\}_{j=1}^K$ to form a complete vector $w^*$.
\end{algorithmic}
\caption{Backsolve: \texttt{BSO-DFO}$(w^0)$}
\label{alg:bs}
\end{algorithm}

\subsubsection{The multi-stage procedure}\label{subsec:multistage}

In this section, we present a multi-stage procedure that iteratively updates and solves local quadratic models, utilizing the \texttt{BSO-DFO} method introduced earlier as the inner solver. 

Our multi-stage procedure employs a scheduler to gradually decrease the NNZ and FLOPs budget, taking small steps towards increased sparsity and reduced FLOPs at every stage. This cautious approach helps maintain the validity of the local quadratic approximation, ensuring that each step is based on accurate information.

The multi-stage method leverages the efficiency of the \texttt{BSO-DFO} algorithm to achieve a balance between computational efficiency and model accuracy. By iteratively solving precise approximations of the true loss function, our approach can efficiently produce pruned networks with superior performance compared to single-stage techniques. The details of the proposed multi-stage procedure can be found in Algorithm \ref{alg:multi}.

\begin{algorithm}[h]
\begin{algorithmic}[1]
\REQUIRE Target NNZ budget $S$ and FLOPs budget $F$, and the number of stages $T_0$.
\STATE Set $w^0=\bar w$; construct sequences of parameters with decreasing NNZ and FLOPs budgets as follows:
\begin{equation}
    S_1 \ge S_2\ge\cdots\ge S_{T_0}=S;\,\,\,F_1 \ge F_2\ge\cdots\ge F_{T_0}=F
\end{equation} 
\FOR {$t=1,2,\ldots,T_0$}
\STATE At current solution $w^{t-1}$, calculate the gradient based on a batch of $n$ data points 
\STATE Construct the objective $Q_L(w)$ as in \eqref{eq:miqp2}, using Hessian approximation \eqref{eq:hessfree} and gradient approximation \eqref{eq:grad-approx}.
\STATE Obtain a solution $w^t$ to problem~\eqref{eq:miqp} with sparsity budget $S_t$ and FLOPs budget $F_t$ by invoking \texttt{BSO-DFO}$(w^{t-1})$. 
\ENDFOR
\end{algorithmic}
\caption{A multi-stage procedure for pruning networks under both NNZ and FLOPs budgets}
\label{alg:multi}
\end{algorithm}

\section{Experimental details}

\subsection{Experimental setup}\label{subsec:expt-setup}
\subsubsection{One-shot pruning experiments}\label{subsubsec:expt-one-shot}
All experiments for one-shot pruning were carried on a computing cluster. Experiments were run on an Intel Xeon Gold 6248 machine with 40 CPUs and one GPU.

\paragraph{Algorithmic setting and hyper-parameters for \modelname~(single-stage)}~\smallskip \\
We apply the \texttt{BSO-DFO} algorithm (Algorithm \ref{alg:bs}) to prune ResNet20, MobileNetV1, and ResNet50. The parameters of \texttt{BSO-DFO} are set as follows: $\tau^s=10^{-3}$, $T=1$ and $\mathcal{Z}_0=\supp(P_{2S,2F}(\bar w))$. For ResNet20 and MobileNetV1, we use $n=1000$ samples for Hessian approximation \eqref{eq:fisherhess} and gradient estimation \eqref{eq:grad-approx}; for ResNet50, we use $n=500$ samples. Each block of the block-wise Hessian approximation has size $B_{size}=2000$. 
We run our proposed methods with a scaling factor $\rho$ of one and a ridge value $\lambda$ ranging from $10^{-6}$ to $10^{-2}$ for each network and FLOPs budget. All results are averaged over 5 runs, and the highest accuracy results are shown.

\paragraph{Algorithmic setting and Hyper-parameters for \modelname++~(multi-stage)}~\smallskip \\
We apply (Algorithm \ref{alg:multi} with the \texttt{BSO-DFO} algorithm as an inner solver to prune ResNet20, MobileNetV1, and ResNet50. The number of stages $T_0$ is set to be $20$. We run our proposed methods with a scaling factor $\rho$ range from $10^2$ to $10^4$ and a ridge value $\lambda$ ranging from $10^{-5}\rho$ to $10^{-2}\rho$ for each network and FLOPs budget. All results are averaged over 5 runs, and the highest accuracy results are shown. The remaining parameters are the same as those in our single-stage method \modelname.

\subsubsection{Gradual pruning experiments}\label{subsubsec:expt-setup-gradual pruning}

All experiments for one-shot pruning were carried on a computing cluster. Experiments for MobileNetV1 were run on an Intel Xeon Platinum 6248 machine with 48 CPUs and 4 GPUs; experiments for ResNet50 were run on four Intel Xeon Platinum 6248 machines with a total of 160 CPUs and 4 GPUs. 

In our gradual pruning experiments, we performed 100 epochs of SGD training with intermittent pruning steps. We employed the polynomial schedule introduced by \citet{ZhuG18} as the pruning schedule to gradually reduce the NNZ and FLOPs budget in each pruning step until reaching our target budget. We utilize the \texttt{BSO-DFO} algorithm as the pruning method.

The learning rate for each epoch $e$ between two pruning steps that occur at epochs $e_1$ and $e_2$ is defined as:
\begin{equation}\label{eq:lr_schedule}
    \textit{min\_lr} + 0.5 \times (\textit{max\_lr} - \textit{min\_lr}) \left( 1 + \cos\left(\pi \frac{e - e_1}{e_2-e_1}\right) \right)
\end{equation}
We used a momentum of 0.9 and a weight decay penalty of ~$3.75\times 10^{-5}$ (taken from ~\citet{Kusupati2020STR}).  

For MobileNetV1, we pruned the network 8 times during training, with 12 epochs between pruning steps. For the first 84 epochs, we set $\textit{min\_lr} = 10^{-5}$ and $\textit{max\_lr} = 0.1$. For remaining epochs, we set $\textit{min\_lr} = 10^{-5}$ and $\textit{max\_lr} = 0.05$. The batch size was set to $4 \times 256$. 

For ResNet50, we pruned the network 7 times during training, with 12 epochs between pruning steps. The $\textit{min\_lr}$ and $\textit{max\_lr}$ values were kept constant as $10^{-5}$ and $0.1$, respectively. The batch size was set to $8 \times 256$.


\subsection{Ablation studies and additional results}\label{subsec:addexp}

\subsubsection{Generalized magnitude pruning}

In Section \ref{sect:impact}, we extend the magnitude pruning approach to cater to both NNZ and FLOP constraints. This part is dedicated to investigating the performance of networks pruned using our generalized magnitude pruning approach, denoted as MP-FLOPs. Table~\ref{tab:accuracy2}  analyzes of the test accuracy between MP-FLOPs, MP, and \modelname~across three networks: ResNet20, MobileNetV1, and ResNet50. As shown in the table, MP-FLOPs consistently perform better than MP in all the tested scenarios. This superior performance results from taking into account both FLOPs and sparsity constraints, which allows MP-FLOPs to consider models with varying FLOPs, NNZ, and accuracy trade-offs, leading to enhanced performance. On the other hand, there remains a significant gap in performance between MP-FLOPs and our proposed model \modelname. This is because our proposed optimization framework considers gradient and Hessian information of the loss function, which allows us to identify a better support of the weight vector refine the weights on the support. This experiment validates the effectiveness of our proposed optimization model \modelname.

\begin{table}[tbp]
    \centering
\caption{The pruning performance (accuracy) of various methods on ResNet20, MobileNetV1, and ResNet50. The bracketed number in the FLOPs column indicates the proportion of FLOPs needed for inference in the pruned network versus the dense network.} 
\label{tab:accuracy2}
    \resizebox{0.6\columnwidth}{!}
    {\begin{tabular}{c|c|ccc}
    \toprule
\footnotesize Network & FLOPs & MP  & MP-FLOPs &  \modelname  \\
\midrule
\multirow{6}{*}{
\begin{minipage}{2cm}
\begin{center}
    ResNet20\\
    on CIFAR10\\
    (91.36\%)
\end{center}
\end{minipage}
}& 24.3M\,(60\%) &88.89 &89.49 &91.38\\ 
& 20.3M\,(50\%) &85.95 &88.36 &90.87\\ 
& 16.2M\,(40\%) &80.42 &84.53 &89.67\\ 
& 12.2M\,(30\%) &58.78 &71.07 &84.42\\ 
& 8.1M\,(20\%) &15.04 &27.20 &65.17\\ 
& 4.1M\,(10\%) &10.27 &18.90 &19.14\\ 
\midrule
\multirow{6}{*}{
\begin{minipage}{2cm}
\begin{center}
    MobileNetV1\\
    on ImageNet\\
    (71.95\%)
\end{center}
\end{minipage}
}& 398M\,(70\%) &70.89 &71.03 &71.83\\ 
& 341M\,(60\%) &67.34 &67.34 &71.42\\ 
& 284M\,(50\%) &51.13 &55.20 &70.35\\ 
& 227M\,(40\%) &14.85 &21.36 &67.18\\ 
& 170M\,(30\%) &0.65 &1.53 &58.40\\ 
& 113M\,(20\%) &0.10 &0.14 &25.82\\  
\midrule
\multirow{6}{*}{
\begin{minipage}{2cm}
\begin{center}
    ResNet50\\
    on ImageNet\\
    (77.01\%)
\end{center}
\end{minipage}
}& 2.3G\,(60\%) &75.38 &75.56 &76.86 \\ 
& 2.0G\,(50\%) &70.85 &72.82 &76.44 \\ 
& 1.6G\,(40\%) &62.37 &64.35 &75.28\\ 
& 1.2G\,(30\%) &22.43 &38.35 &71.54\\ 
& 817M\,(20\%) &0.56 &1.55 &54.27\\ 
& 408M\,(10\%) &0.10 &0.11 &0.46\\ 
\bottomrule
 \end{tabular} }
\end{table} 

\subsubsection{Sparsity of pruned models}

In Table \ref{tab:sparsity}, we present the percentages of NNZs for ResNet20, MobileNetV1, and ResNet50 that have been pruned using different methods. By simultaneously accounting for both FLOP and NNZ constraints, \modelname~prunes the network that strikes the optimal balance between FLOPs and NNZs. This leads to networks with slightly more NNZs, yet significantly higher accuracy compared to other approaches.

\begin{table}[!htbp]
    \centering
\caption{The percetage of NNZs in ResNet20, MobileNetV1, and ResNet50 pruned by various methods. The bracketed number in the FLOPs column indicates the proportion of FLOPs needed for inference in the pruned network versus the dense network.  } 
\label{tab:sparsity}
    \resizebox{0.6\columnwidth}{!}
    {\begin{tabular}{c|c|cccc}
    \toprule
\footnotesize Network & FLOPs & MP  & WF & CHITA & \modelname \\
\midrule
\multirow{6}{*}{
\begin{minipage}{2cm}
\begin{center}
    ResNet20\\
    on CIFAR10\\
    (91.36\%)
\end{center}
\end{minipage}
}& 24.3M\,(60\%) &53.9\% &59.3\% &53.9\% &60.0\%\\ 
& 20.3M\,(50\%) &43.1\% &49.1\% &43.1\% &53.0\%\\ 
& 16.2M\,(40\%) &32.6\% &39.1\% &32.6\% &46.0\%\\ 
& 12.2M\,(30\%) &22.7\% &29.1\% &22.7\% &33.0\%\\ 
& 8.1M\,(20\%) &13.5\% &19.1\% &13.5\% &23.0\%\\ 
& 4.1M\,(10\%) &5.2\% &9.0\% &5.2\% &11.0\%\\ 
\midrule
\multirow{6}{*}{
\begin{minipage}{2cm}
\begin{center}
    MobileNetV1\\
    on ImageNet\\
    (71.95\%)
\end{center}
\end{minipage}
}& 398M\,(70\%) &66.3\% &71.1\% &66.3\% &69.0\%\\ 
& 341M\,(60\%) &55.6\% &63.1\% &55.6\% &59.0\%\\ 
& 284M\,(50\%) &44.9\% &54.8\% &44.9\% &52.0\%\\ 
& 227M\,(40\%) &34.1\% &45.9\% &34.1\% &45.0\%\\ 
& 170M\,(30\%) &23.5\% &36.5\% &23.5\% &31.0\%\\ 
& 113M\,(20\%) &13.4\% &26.2\% &13.4\% &21.0\%\\ 
\midrule
\multirow{6}{*}{
\begin{minipage}{2cm}
\begin{center}
    ResNet50\\
    on ImageNet\\
    (77.01\%)
\end{center}
\end{minipage}
}& 2.3G\,(60\%) &56.2\% &64.4\% &56.2\% &65.0\%\\ 
& 2.0G\,(50\%) &44.9\% &54.6\% &44.9\% &55.0\%\\ 
& 1.6G\,(40\%) &33.6\% &44.4\% &33.6\% &48.0\%\\ 
& 1.2G\,(30\%) &22.8\% &33.8\% &22.8\% &38.0\%\\ 
& 817M\,(20\%) &12.7\% &22.8\% &12.7\% &25.0\%\\ 
& 408M\,(10\%) &4.4\% &11.4\% &4.4\% &15.0\%\\ 
\bottomrule
 \end{tabular} }
\end{table}

\subsubsection{Additional comparison with other pruning method}

We conduct an additional one-shot pruning comparison between \modelname~and CAIE \cite{wu2020constraint}. Unlike \modelname~and other competitors, CAIE is a structured pruning approach that prunes entire filters in a convolutional network, not just individual weights. Nonetheless, CAIE's methodology is adaptable to unstructured pruning scenarios. For a fair comparison, we implemented CAIE's algorithm in an unstructured pruning setting, assessing CAIE's loss impact using a second-order Taylor approximation and calibrating its NNZ budget to optimize accuracy. The results, presented in Table \ref{tab:caie}, demonstrate that \modelname~consistently outperforms CAIE in terms of accuracy. This superior performance can be attributed to the efficacy of our proposed optimization method.

\begin{table}    
\centering
    \resizebox{0.98\columnwidth}{!}
    {\begin{tabular}{cccc|cccc|cccc}
    \toprule
\footnotesize Network & FLOPs & CAIE & \texttt{FALCON} & Network & FLOPs & CAIE & \texttt{FALCON} &Network & FLOPs & CAIE & \texttt{FALCON}  \\ \midrule 
\multirow{6}{*}{
\begin{minipage}{2cm}
\begin{center}
    ResNet20\\
    on CIFAR10\\
    (91.36\%)
\end{center}
\end{minipage}
}
& 24.3M\,(60\%) &89.54 &91.38 &  \multirow{6}{*}{
\begin{minipage}{2cm}
\begin{center}
    MobileNetV1\\
    on ImageNet\\
    (71.95\%)
\end{center}
\end{minipage}
} & 398M\,(70\%) &70.48 &71.83 & \multirow{6}{*}{
\begin{minipage}{2cm}
\begin{center}
    ResNet50\\
    on ImageNet\\
    (77.01\%)
\end{center}
\end{minipage}
}& 2.3G\,(60\%) &75.51 &76.86 \\
& 20.3M\,(50\%) &87.33 &90.87 & & 341M\,(60\%) &66.72 &71.42 & & 2.0G\,(50\%) &72.98 &76.39\\ 
& 16.2M\,(40\%) &80.83 &89.67 & & 284M\,(50\%) &54.12 &70.35 & & 1.6G\,(40\%) &63.34 &75.28\\ 
& 12.2M\,(30\%) &52.44 &84.42 & & 227M\,(40\%) &17.01 &67.18 & & 1.2G\,(30\%) &29.41 &71.54 \\ 
& 8.1M\,(20\%) &23.18 &65.17  & & 170M\,(30\%) &1.25 &58.40  & & 817M\,(20\%) &0.75 &54.27  \\ 
& 4.1M\,(10\%) &15.13 &19.14 & &  113M\,(20\%) &0.12 &25.82  & & 408M\,(10\%) &0.10 &0.46 \\ 
\bottomrule
 \end{tabular} }
 \vspace{1mm}
 \caption{The accuracy comparison between CAIE and \modelname~on ResNet20, MobileNetV1, and ResNet50.}
 \label{tab:caie}
\end{table}

\subsubsection{Sparsity distribution of models pruned by \modelname} 

In this part, we assess the sparsity distribution of models pruned using \modelname~within a fixed FLOP budget ($F_0$). We explore three distinct scenarios: (i) pure FLOP constraint: we set the FLOP budget to $F=F_0$ and the NNZ budget to $S=\infty$; (ii) pure sparsity constraint: we fix $F=\infty$ and find the NNZ budget $S$ so that the FLOPs of the resulting network precisely equals $F_0$; (iii) joint sparsification: we choose $F=F_0$ and $S$ optimally to maximize accuracy.  

\begin{figure}[!htbp]
   \centering
   \begin{subfigure}[b]{0.7\textwidth}
   \centering
       \includegraphics[width=0.99\columnwidth,trim=2cm 0.5 2cm 0.5cm]{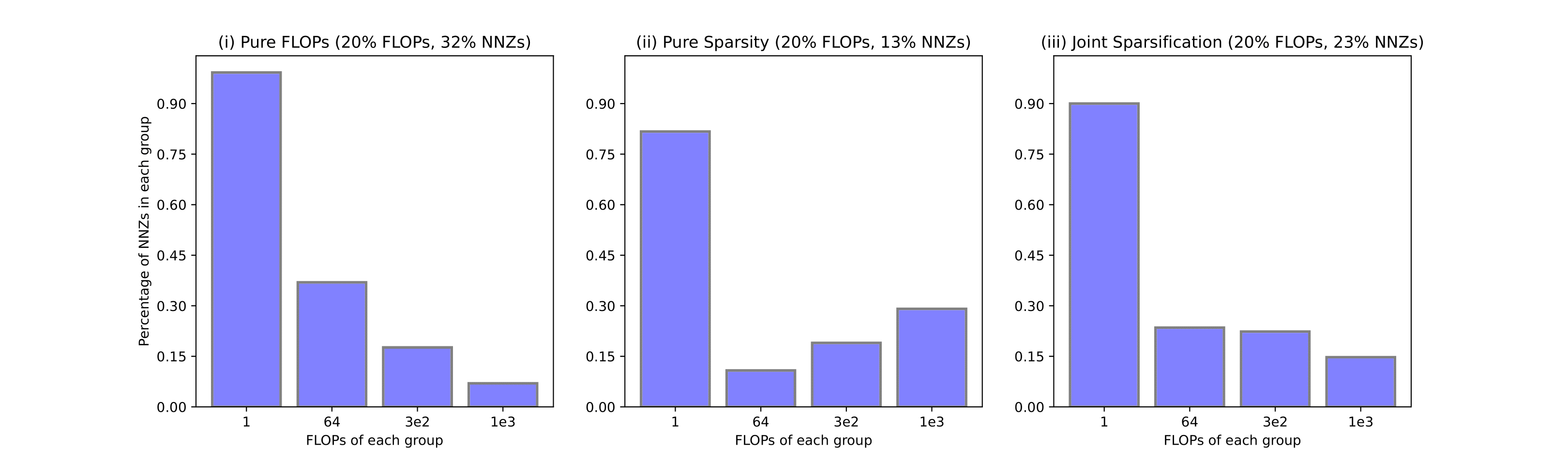}
       \caption{Resnet20}
   \end{subfigure}
   \vspace{2mm}
   \begin{subfigure}[b]{0.7\textwidth}
   \centering
       \includegraphics[width=0.99\columnwidth,trim=2cm 0.5 2cm 0.5cm]{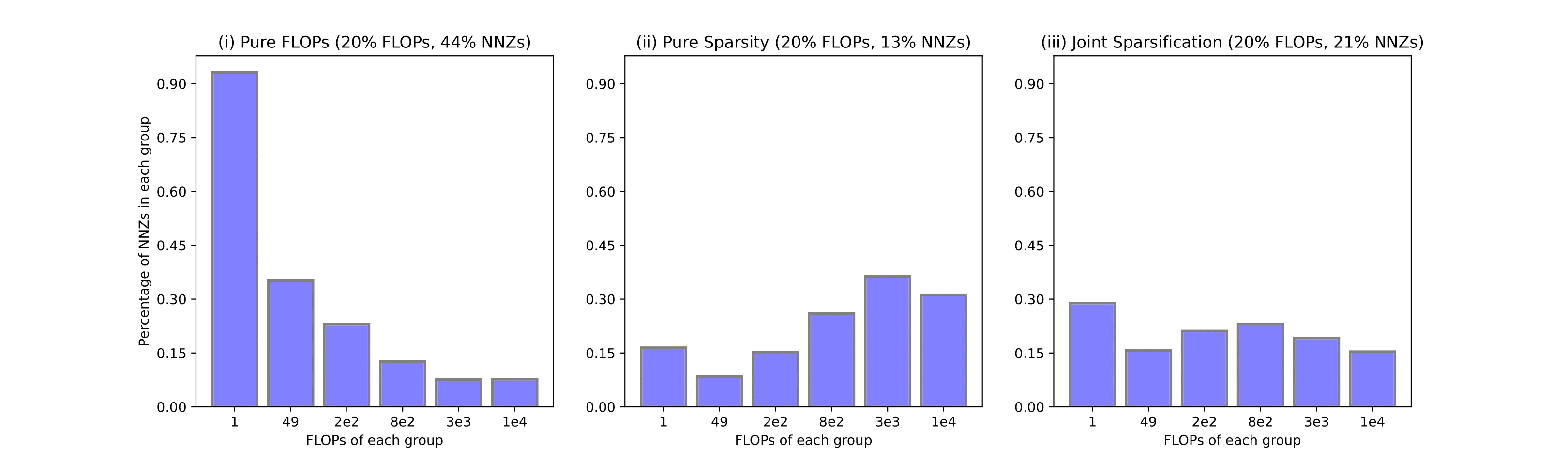}
       \caption{MobileNetV1}
   \end{subfigure}
   \vspace{1mm}
   \begin{subfigure}[b]{0.7\textwidth}
   \centering
       \includegraphics[width=0.99\columnwidth,trim=2cm 0.5 2cm 0.5cm]{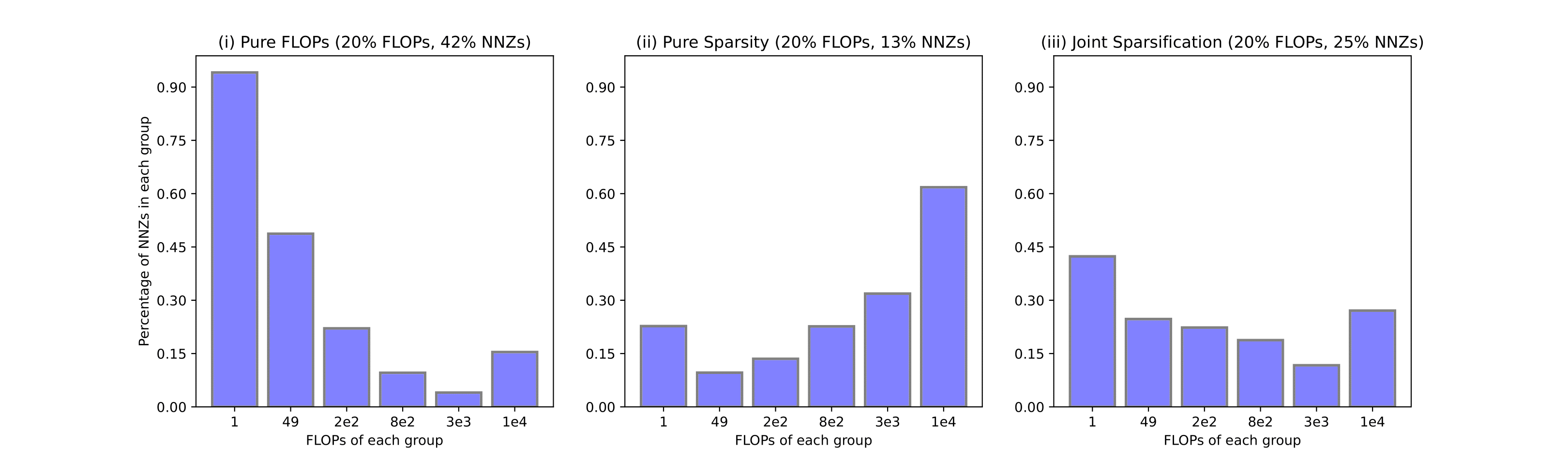}
       \caption{Resnet50}
   \end{subfigure}
     \caption{Sparsity of each group of models pruned under a fixed FLOP budget (20\% of total FLOPs) by \modelname~across three scenarios: pure FLOP constraint, pure sparsity constraint, and joint sparsification.}
     \label{fig:bar2}
\end{figure}

In Figure \ref{fig:bar2}, we display the group-wise sparsity in Resnet20, MobileNetV1, and Resnet50 models pruned by \modelname~in these three scenarios. Our findings confirm that joint sparsification achieves more balanced sparsity across groups. This substantiates the claim in Section \ref{subsec:oneshot}. And as explained in Section \ref{subsec:oneshot}, this is why joint sparsification achieves higher accuracy than pure FLOPs or sparsity pruning.

\end{document}